\documentclass[11pt]{article}

\usepackage[final]{acl}

\usepackage{times}
\usepackage{latexsym}

\usepackage[T1]{fontenc}

\usepackage[utf8]{inputenc}

\usepackage{microtype}

\usepackage{inconsolata}

\usepackage{graphicx}
\usepackage{marvosym}
\newcommand\blfootnote[1]{%
  \begingroup
  \renewcommand\thefootnote{}\footnote{#1}%
  \addtocounter{footnote}{-1}%
  \endgroup
}
\usepackage{listings}
\usepackage[table]{xcolor}
\usepackage[most]{tcolorbox}
\tcbuselibrary{listings,breakable}

\definecolor{PromptBack}{RGB}{252,248,232}    
\definecolor{PromptTitle}{RGB}{232,224,188}   
\definecolor{PromptFrame}{RGB}{185,170,120}   

\lstdefinestyle{promptstyle}{
  basicstyle=\ttfamily\footnotesize,
  breaklines=true,
  breakatwhitespace=true,
  breakautoindent=false,
  breakindent=0pt,
  columns=fullflexible,
  keepspaces=false,
  showstringspaces=false,
  resetmargins=true
}

\newtcblisting{promptbox}[1]{
  enhanced,
  breakable,
  listing only,
  listing options={style=promptstyle},
  colback=PromptBack,
  colframe=PromptFrame,
  colbacktitle=PromptTitle,
  coltitle=black,
  title={#1},
  fonttitle=\small\bfseries,
  boxrule=0.45pt,
  arc=1pt,
  left=1mm,
  right=1mm,
  top=1mm,
  bottom=1mm,
  before skip=5pt,
  after skip=5pt
}

\usepackage{tabularx}
\usepackage{array}

\newcolumntype{Y}{>{\raggedright\arraybackslash}X}
\newcolumntype{C}[1]{>{\centering\arraybackslash}p{#1}}
\newcolumntype{L}[1]{>{\raggedright\arraybackslash}p{#1}}

\author{\textbf{Siyuan Huang}$^{*1}$, \textbf{Yifan Zhou}$^{*2}$, \textbf{Yutong Gao}$^{*3}$, \textbf{Zi Yin}$^{*4}$, \textbf{Juyang Bai}$^{*1}$, \\ \textbf{Xinxin Liu}$^{5}$, \textbf{Rama Chellappa}$^{1}$, \textbf{Chun Pong Lau}$^{6}$, \textbf{Cheng Peng}$^{1}$, \\ \textbf{Sayan Nag}$^{7}$, \textbf{Shraman Pramanick}$^{7}$\textsuperscript{\Letter} \vspace{2mm}\\
$^{1}$Johns Hopkins University \ \ $^{2}$Zhejiang University \ \ $^{3}$Independant Researcher \\ $^{4}$Tsinghua University \ \ $^{5}$University of Central Florida \ \ $^{6}$City University of Hong Kong \\ $^{7}$Adobe Research \vspace{2mm}
\\
}

\usepackage{graphicx}
\usepackage{booktabs}
\usepackage{multirow}
\usepackage{multicol}
\usepackage{caption}
\usepackage{amsmath}
\usepackage{amssymb}

\usepackage[table]{xcolor}

\hypersetup{
  colorlinks=true,
  linkcolor=blue,
  citecolor=blue,
  urlcolor=magenta
}

\definecolor{Light}{HTML}{f7ffd4}
\newcommand{\method}{\text{SciFig}}
\newcommand{\eval}{\text{SciFig-Eval}}
\newcommand{\data}{\text{SciFig-Bench}}

\title{\method: Towards Automating Editable Figure \\ Generation for Scientific Papers}

\begin{document}

\maketitle

\begin{abstract}
High-quality methodology figures are central to scientific communication, yet they remain difficult and time-consuming to create. Such figures must distill a method's components and information flow into a clear, revisable diagram as the paper evolves. Existing methodology diagram automation systems typically face a trade-off between editability and visual quality: TikZ- or SVG-based methods produce editable structured outputs but often lack the richness of human-designed figures, while image-generation models produce polished raster outputs that are difficult to revise. We introduce \method{}, an end-to-end multi-agent framework for generating visually rich and fully editable methodology figures from scientific text. \method{} decomposes figure generation into planning, layout synthesis, component rendering, and iterative refinement, producing XML figures that can be edited in standard diagramming tools and refined through human or VLM feedback. We also introduce \data{}, a human-verified benchmark of 435 author-drawn methodology figures from 37 arXiv domains and 15 top-tier AI/ML venues, and \eval{}, a four-axis evaluation protocol for measuring figure quality. Across seven single-agent and agentic baselines, \method{} achieves the best performance on all four \eval{} axes and generates editable figures in about 10 minutes on average. Qualitative examples further show that \method{} can generalize beyond methodology figures to teaser diagrams and statistical plots. Dataset and code are available at: \url{https://shramanpramanick.github.io/SciFig/}.
\end{abstract}
\blfootnote{\textsuperscript{*}equal technical contribution.}
\blfootnote{\textsuperscript{\Letter}lead, corresponding author: \url{spramanick@adobe.com}.}

\vspace{-3mm}
\section{Introduction}
\label{sec:intro}
\vspace{-1mm}

Vision-language models and multimodal generative models are reshaping the scientific research lifecycle, from literature review~\cite{wangautosurvey, wu2025automatedliterature, wu2025autosurvey2, asai2026synthesizing} and idea generation~\cite{gottweis2025towardsanaicoscientist, baek2025researchagent, su2025manyheads, ai-researcher} to experiment iteration~\cite{huang2024mlagentbench, yamada2025aiscientistv2, novikov2025alphaevolve} and paper writing~\cite{wengcycleresearcher, wang2025scholarcopilot, song2026paperorchestra}. Yet visual communication, particularly scientific figure creation, remains largely manual. While statistical plots can often be generated using programming tools\footnote{Examples include \href{https://matplotlib.org/}{Matplotlib}, \href{https://seaborn.pydata.org/}{Seaborn}, and \href{https://ggplot2.tidyverse.org/}{ggplot2}.}, methodology figures pose a distinct challenge: they must communicate a system's structure, components, roles, and information flow, allowing readers to quickly grasp a paper's central idea. Creating such figures requires interpreting unstructured technical prose, identifying the method's logical organization, planning a spatial layout, and rendering coherent visual components consistent with scientific publishing conventions~\cite{schmied2021effective}.

\begin{figure}[!t]
\centering
\includegraphics[width=0.46\textwidth]{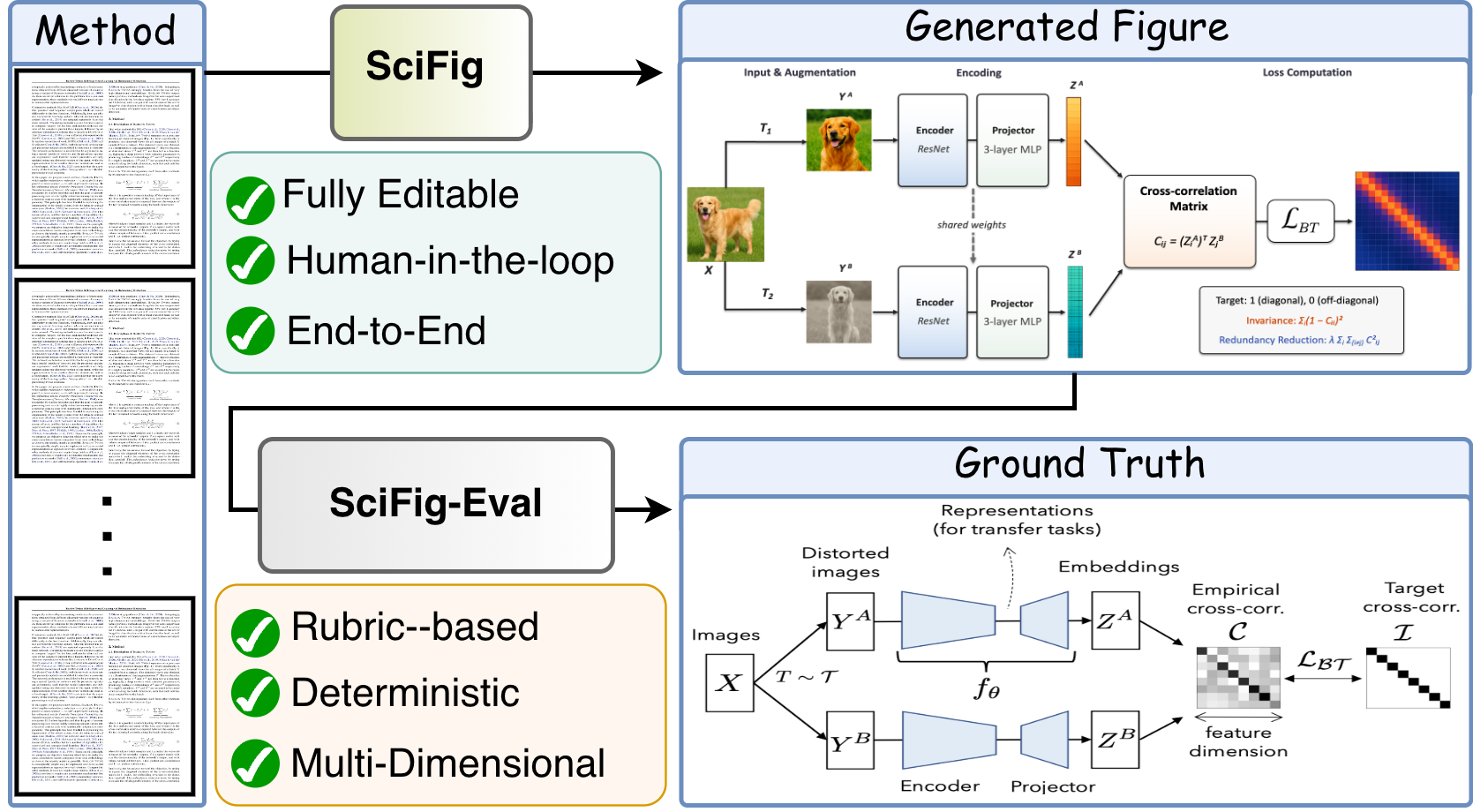}
\vspace{-1mm}
\caption{\textbf{\method{}} generates visually rich, fully editable methodology figures from method text and figure captions; \textbf{\eval{}} evaluates them against author-drawn references with a rubric-based, deterministic, multi-dimensional protocol. Each figure remains end-to-end editable and supports human-in-the-loop refinement. Running example: Barlow Twins~\cite{zbontar2021barlowtwins}.}
\vspace{-5mm}
\label{fig:teaser}
\end{figure}

Recent work has begun to address this problem using LLMs and VLMs to generate scientific illustrations from text. One line of work produces structured, editable representations such as TikZ or SVG~\cite{belouadiautomatikz, guo2025paper2sysarch, mondal2024scidoc2diagrammer, lin2026autofigureedit}. However, because these representations rely on structured vector primitives and symbolic drawing commands, they often struggle to capture the texture, shading, visual density, and stylistic variation of RGB illustrations. Another line leverages VLMs and image-generation models to produce visually expressive, publication-ready figures~\cite{zhu2026autofigure, zhu2026paperbanana}, but typically outputs raster images that are difficult to edit and often require full regeneration for small revisions. Despite their different output forms, both directions still struggle to faithfully translate long, unstructured methodology text into figures that preserve fine-grained components, module interactions, and information flow.

In this work, we introduce \method{}, an end-to-end multi-agent framework for generating methodology figures that are both visually rich and fully editable. \method{} bridges raster and symbolic approaches by combining rendered visual components with styled text elements, while exporting XML figures whose elements can be independently edited in standard diagramming tools\footnote{Such as \href{https://www.microsoft.com/en-us/microsoft-365/powerpoint}{PowerPoint}, \href{https://www.apple.com/keynote/}{Keynote}, \href{https://www.adobe.com/products/illustrator.html}{Illustrator}, \href{https://www.drawio.com/}{diagrams.net}.}. It decomposes figure generation into four stages: planning, layout synthesis, component rendering, and iterative feedback, each handled by a dedicated agent (Section~\ref{sec:system}). This design separates structural layout from visual rendering, enabling image-level expressiveness without sacrificing editability, while grounding refinement in the full methodology text to preserve components, interactions, and information flow. \method{} supports both \emph{human-in-the-loop} refinement through direct edits or natural-language feedback and \emph{VLM-in-the-loop} refinement through automatic visual critiques. Figure~\ref{fig:teaser} summarizes this generation and evaluation setting. Although designed and benchmarked primarily for method figures, \method{} can also generate teaser figures and statistical plots, as shown in qualitative examples.

To support systematic benchmarking, we introduce \data{}, a human-verified benchmark of 435 methodology figures with source text and captions spanning 37 AI/ML-centered domains and 15 top-tier venues, along with a separate set of 1,784 figures for developing evaluation rubrics (Section~\ref{sec:data_curation}). Beyond generation, reliably evaluating methodology figures also remains challenging: a good figure must preserve key components, relationships, and information flow while allowing multiple valid layouts. Existing evaluation in automatic poster~\cite{paper2poster, zhang2025postergen}, figure~\cite{guo2025paper2sysarch, zhu2026paperbanana, zhu2026autofigure}, slide~\cite{zeng2025slidetailor, liang2025slidegen, zheng2025pptagent, ge2025autopresent}, and webpage~\cite{chen2025paper2web} generation typically reduces this multidimensional problem to a single VLM-judged numerical score, which conflates failure modes and is sensitive to model choice, scale range, and prompt phrasing. We therefore introduce \eval{}, a decomposed evaluation protocol with four axes: \emph{Completeness and Correctness} via cross-model question answering, \emph{Rubric-Based Content Quality} via 20 binary questions derived from real figures, \emph{Perceptual Design Quality} via deterministic heuristic metrics, and \emph{Reference-Based Fidelity} via visual, cross-modal, and textual comparison against the author-drawn figure. By combining binary judgments, deterministic heuristics, reference-based metrics, and multi-model aggregation, \eval{} provides a more reproducible and diagnostic alternative to single-score VLM judging.

Overall, our contributions are three-fold:


\noindent \textbf{\method{}}: We design an end-to-end multi-agent framework for generating visually rich and fully editable methodology figures from scientific text. \method{} supports human-in-the-loop and VLM-in-the-loop refinement and outperforms evaluated single- and multi-agent baselines, including editable and non-editable systems. In a human study with 60 participants, \method{} achieves the strongest pairwise preference while generating editable figures in under 10 minutes on average.

\noindent \textbf{\data{}}: We curate a human-verified benchmark of 435 methodology figures spanning 37 AI/ML-centered domains across 15 top-tier venues.

\noindent \textbf{\eval{}}: We propose a multi-dimensional evaluation framework that assesses figure quality along four complementary axes and is validated against judgments from 60 experts, achieving Pearson correlation up to $r=0.92$ ($p < 0.001$).

\vspace{-2mm}
\section{Related Works}
\vspace{-2mm}

\begin{figure*}[!t]
\centering
\includegraphics[width=0.98\textwidth]{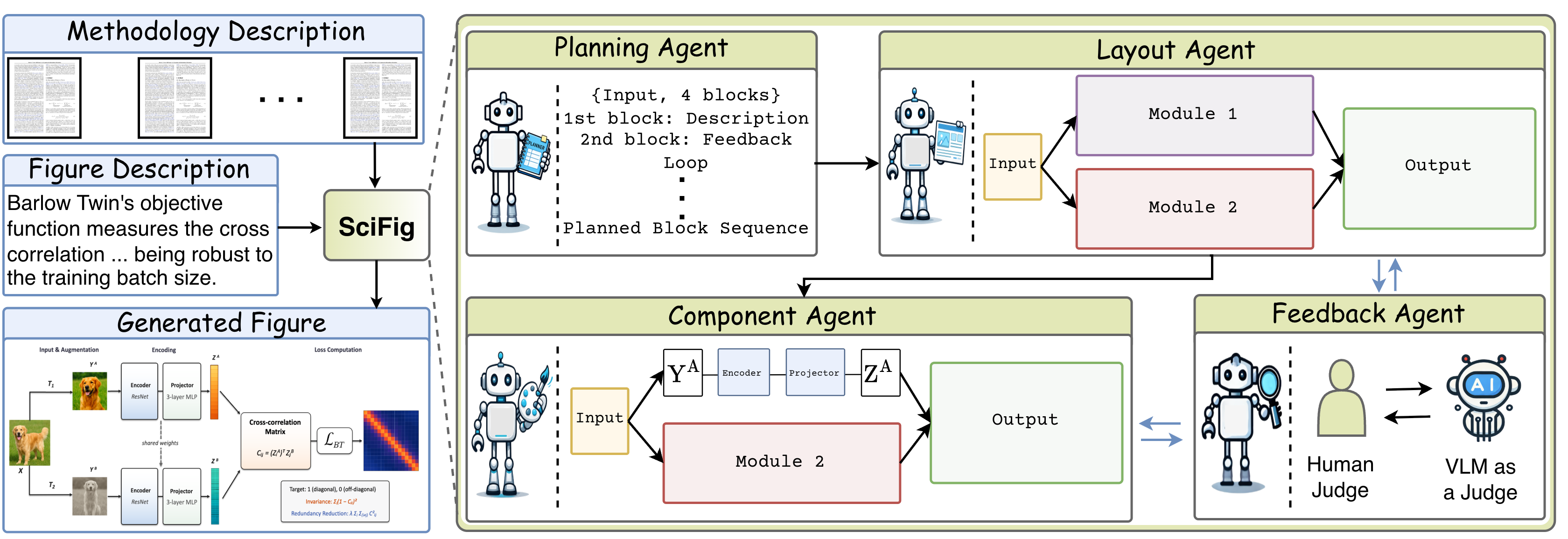}
\vspace{-2mm}
\caption{\textbf{\method{} generates editable methodology figures from text through four cooperating agents.} Given method text and a figure caption, the \textbf{Planning Agent} extracts pipeline blocks; the \textbf{Layout Agent} converts the plan into editable XML; the \textbf{Component Agent} adds visual elements; and the \textbf{Feedback Agent} refines the figure using human or VLM critiques. This figure was generated by \method{} and refined through the human-in-the-loop interface.}
\vspace{-5mm}
\label{fig:method}
\end{figure*}

\noindent \textbf{LLMs and VLMs in Scientific Research Workflows.}
Recent work has applied LLMs and VLMs across the scientific research pipeline, including literature review and survey generation~\cite{wangautosurvey, agarwallitllms, wu2025automatedliterature, wu2025autosurvey2, nguye2025surveyg, asai2026synthesizing, lala2023paperqa, taylor2022galactica, shao2024storm}, hypothesis and idea generation~\cite{gottweis2025towardsanaicoscientist, baek2025researchagent, su2025manyheads, ai-researcher, wang2023scimon}, closed-loop experimentation~\cite{lu2024aiscientist, yamada2025aiscientistv2, huang2024mlagentbench, novikov2025alphaevolve, bran2023chemcrow, ghafarollahi2024sciagents}, and paper writing~\cite{wengcycleresearcher, wang2025scholarcopilot, song2026paperorchestra}. While these systems synthesize prior work, propose research directions, run experiments, or draft manuscripts, \method{} addresses a comparatively under-automated artifact in scientific communication: the methodology figure.

\vspace{1mm}
\noindent \textbf{Automatic Generation of Scientific Figures, Posters, and Slides.}
Prior systems generate scientific figures through structured editable representations, including TikZ~\cite{belouadiautomatikz, belouadi2024detikzify}, SVG-based layouts~\cite{guo2025paper2sysarch, mondal2024scidoc2diagrammer, lin2026autofigureedit, zala2023diagrammergpt}, and raw vector code~\cite{rodriguez2024starvector}, or through raster outputs from VLMs and image-generation models~\cite{zhu2026autofigure, zhu2026paperbanana}. Related work also targets statistical charts and plots~\cite{han2023chartllama, wu2024plot2code, yang2024matplotagent}, multi-page posters~\cite{paper2poster, zhang2025postergen}, slide decks~\cite{zheng2025pptagent, ge2025autopresent, zeng2025slidetailor, liang2025slidegen}, and webpage-style layouts~\cite{chen2025paper2web}. In contrast, \method{} focuses on methodology figures, jointly addressing visual richness, full editability, and faithfulness to long, technical method descriptions.

\vspace{1mm}
\noindent \textbf{Evaluation of Generated Scientific Content.}
Evaluating generated scientific content remains challenging. Most prior work prompts a single VLM for a holistic numerical score~\cite{paper2poster, zhang2025postergen, zhu2026autofigure, zhu2026paperbanana, zheng2025pptagent, ge2025autopresent, chen2025paper2web}, which is sensitive to model choice, scale range, and prompt phrasing~\cite{zheng2023judging, wang2023largelm, liu2023geval}. Question-based protocols such as PaperQuiz~\cite{paper2poster} and visual question-answering benchmarks~\cite{masry2022chartqa, kahou2017figureqa} use binary or multiple-choice probes, while recent benchmarks study end-to-end scientific reproduction~\cite{starace2025paperbench}. \eval{} extends these directions by decomposing methodology-figure quality into four axes that combine cross-model VQA, rubric-based binary judgments from real figures, deterministic perceptual heuristics, and reference-based metrics, providing a more reproducible and diagnostic alternative to single-score VLM judging.

\vspace{-2mm}
\section{Method}
\vspace{-2mm}

\subsection{Problem Formulation} \label{sec:problem_formulation}

We formalize methodology figure generation as mapping scientific text to an editable visual representation. Given methodology description $T$ (e.g., a paper's method section) and figure description $D$ (e.g., a caption or high-level visual specification), the goal is to generate a figure $F=f(T,D)$ that faithfully depicts method $T$ while following $D$.

Methodology figures are \emph{structured compositions} of visual components---blocks, arrows, labels, and sub-modules---arranged to reflect the method's logical organization and information flow. We represent a figure as $F=(V,L,E)$, where $V=\{v_1,\ldots,v_m\}$ denotes visual components, $L=\{(p_i,s_i)\}_{i=1}^m$ specifies their positions and sizes, and $E\subseteq V\times V$ defines directed connections.

\vspace{1mm}
\noindent \textbf{Editability.} A key requirement in our setting is \emph{editability}: researchers often revise figures for venue formatting, reviewer feedback, or architectural changes. We define a system as editable if: (1) individual components $v_i$ can be repositioned, resized, or restyled without regenerating the full figure, and (2) the output is a directly manipulable vector format, such as XML. Image-generation models such as GPT-Image~\cite{gpt5.2}, Qwen-Image~\cite{wu2025qwenimage}, and Nano Banana Pro~\cite{nanobananapro} can produce visually appealing figures, but their raster outputs are difficult to edit: even small changes require regeneration and may alter unchanged regions. In contrast, \method{} produces structured editable code, supporting both \emph{human-in-the-loop} refinement, where researchers edit the code or rendered output, and \emph{VLM-in-the-loop} refinement, where a VLM inspects the rendered figure and suggests revisions.

\vspace{-2mm}
\subsection{\method: System Architecture} \label{sec:system}

\method{} decomposes figure generation into four stages: planning, layout synthesis, component rendering, and iterative feedback. As shown in Figure~\ref{fig:method}, the system outputs an editable XML file encoding $F=(V,L,E)$. We use Barlow Twins~\cite{zbontar2021barlowtwins} as the running example in Figure~\ref{fig:method} and throughout this section.

\vspace{1mm}
\noindent \textbf{Planning agent.} Given $T$ and $D$, the planning agent produces a textual plan $P=\mathcal{A}_{\text{plan}}(T,D)$ specifying the layout direction, aspect ratio, pipeline stages, components, functional groupings, and internal connections. For Barlow Twins, it identifies a horizontal siamese structure with two parallel branches, each containing data augmentation, a shared encoder, and a projector MLP, followed by loss computation. This separates \emph{what} should be depicted from \emph{where} elements should appear.

\vspace{1mm}
\noindent \textbf{Layout agent.} The layout agent converts $P$ into an initial XML layout $L_0=\mathcal{A}_{\text{layout}}(P)$, encoding positions $p_i$, sizes $s_i$, and connections $E$. The layout has three levels: modules for pipeline stages, sub-groups for related operations within a module, and leaf components for individual operations. Thick arrows denote main inter-module flow, while thin arrows capture intra-module connections. Because the layout is structured markup rather than an image, it directly supports the editability requirement in Section~\ref{sec:problem_formulation}.

\vspace{1mm}
\noindent \textbf{Component agent.} The component agent generates visual elements $V=\{v_1,\ldots,v_m\}$ to populate the layout. Each leaf component is rendered in either \emph{PNG mode}, where an image-generation model creates a visually rich element embedded in XML as a base64 image, or \emph{textbox mode}, where a styled labeled rectangle is produced natively in draw.io. The plan $P$ guides this choice: visually meaningful components use PNG mode, while named operations and self-explanatory variables use textbox mode. For Barlow Twins, the cross-correlation heatmap uses PNG mode, while variables such as $Y^{A}$, $Y^{B}$, $Z^{A}$, and $Z^{B}$ use textbox mode.

\vspace{1mm}
\noindent \textbf{Feedback agent.} The feedback agent refines both layout and components. In \emph{human-in-the-loop} mode, researchers directly edit the figure or provide natural-language instructions, such as ``align the two siamese branches vertically'' or ``replace the encoder icon with a ResNet diagram.'' In \emph{VLM-in-the-loop} mode, a VLM critiques the rendered figure and suggests revisions to improve layout, alignment, readability, and component quality. The two modes can be alternated across rounds. During layout refinement, the feedback agent runs $K$ rounds:
\begin{gather}
\vspace{-2mm}
\text{FB}_k = \mathcal{A}_{\text{feedback}}\!\big(\texttt{render}(L_{k-1}),\; T\big), \nonumber \\
L_k = \mathcal{A}_{\text{layout}}\!\big(P,\; L_{k-1},\; \text{FB}_k\big).
\vspace{-2mm}
\end{gather}
After the component agent populates the stabilized layout $L^*$, the feedback agent similarly refines the composed figure $F=\Phi(L^*,V)$, correcting issues that emerge after component embedding, such as label-component overlaps, sizing mismatches, or unclear visual elements. Additional implementation details are provided in the appendix \ref{appendix:impl}.

\vspace{-1mm}
\section{Benchmark \& Evaluation Setup}
\vspace{-1mm}

\subsection{\data: Data Curation}
\label{sec:data_curation}

To evaluate \method{} across diverse pipeline figures, we construct a filtered, human-verified benchmark where each sample contains: ($i$) methodology text describing the proposed system, ($ii$) the corresponding author-drawn figure, and ($iii$) its caption.

\begin{figure*}[!t]
\centering
\begin{minipage}[t]{0.54\linewidth}
\centering
\includegraphics[width=\linewidth]{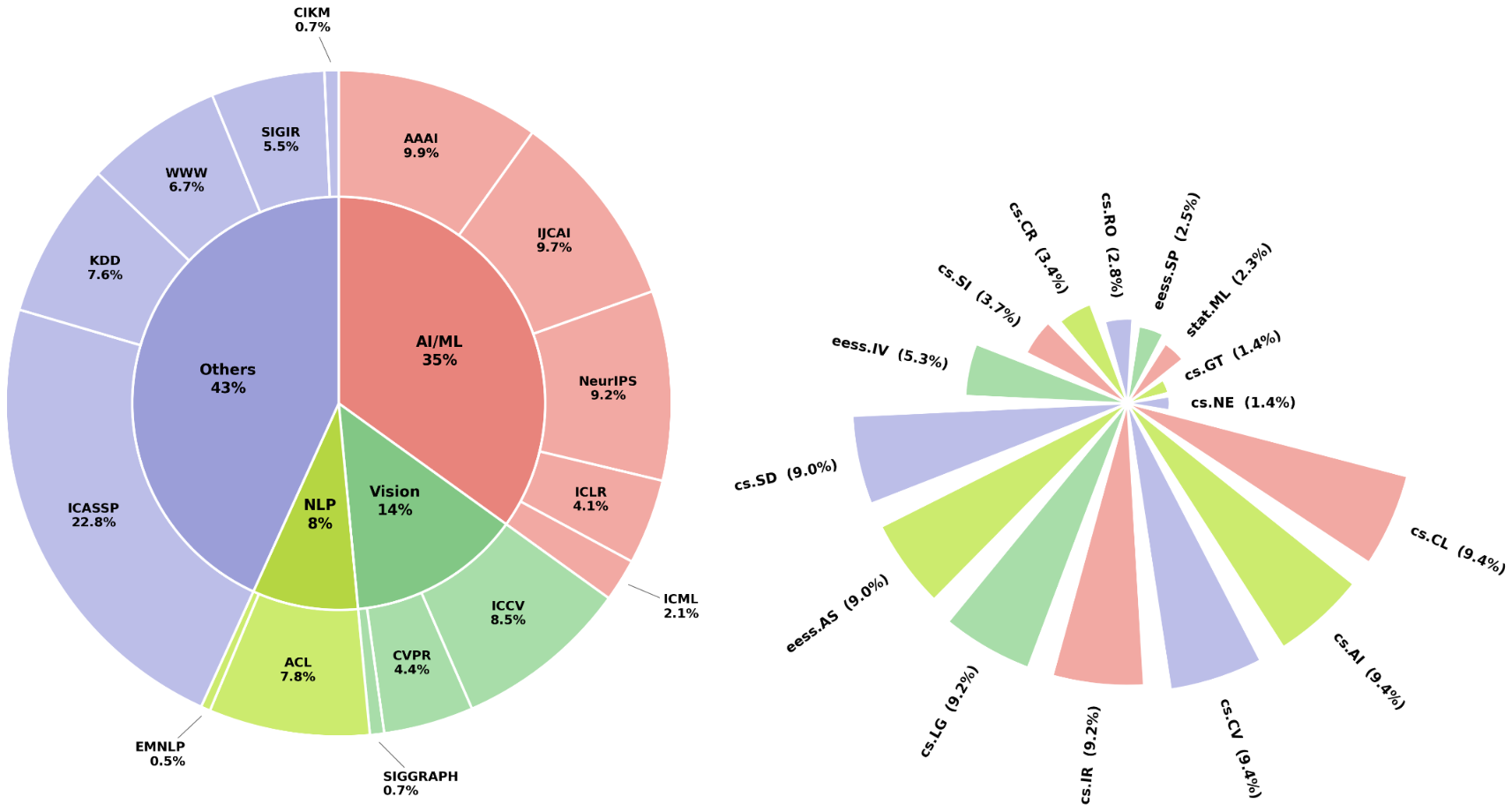}
\caption{\textbf{Venue and domain distribution of \data{}.} Left:
two-level pie chart of the 15 top-tier AI/ML conference venues
represented in our benchmark, including NeurIPS, ICML, CVPR,
ACL, EMNLP etc. Right: radial chart of arXiv subject categories across the 435 papers, spanning 37 domains.}
\label{fig:paper_domains}
\end{minipage}\hfill
\hspace{0.5mm}
\begin{minipage}[t]{0.435\linewidth}
\centering
\includegraphics[width=\linewidth]{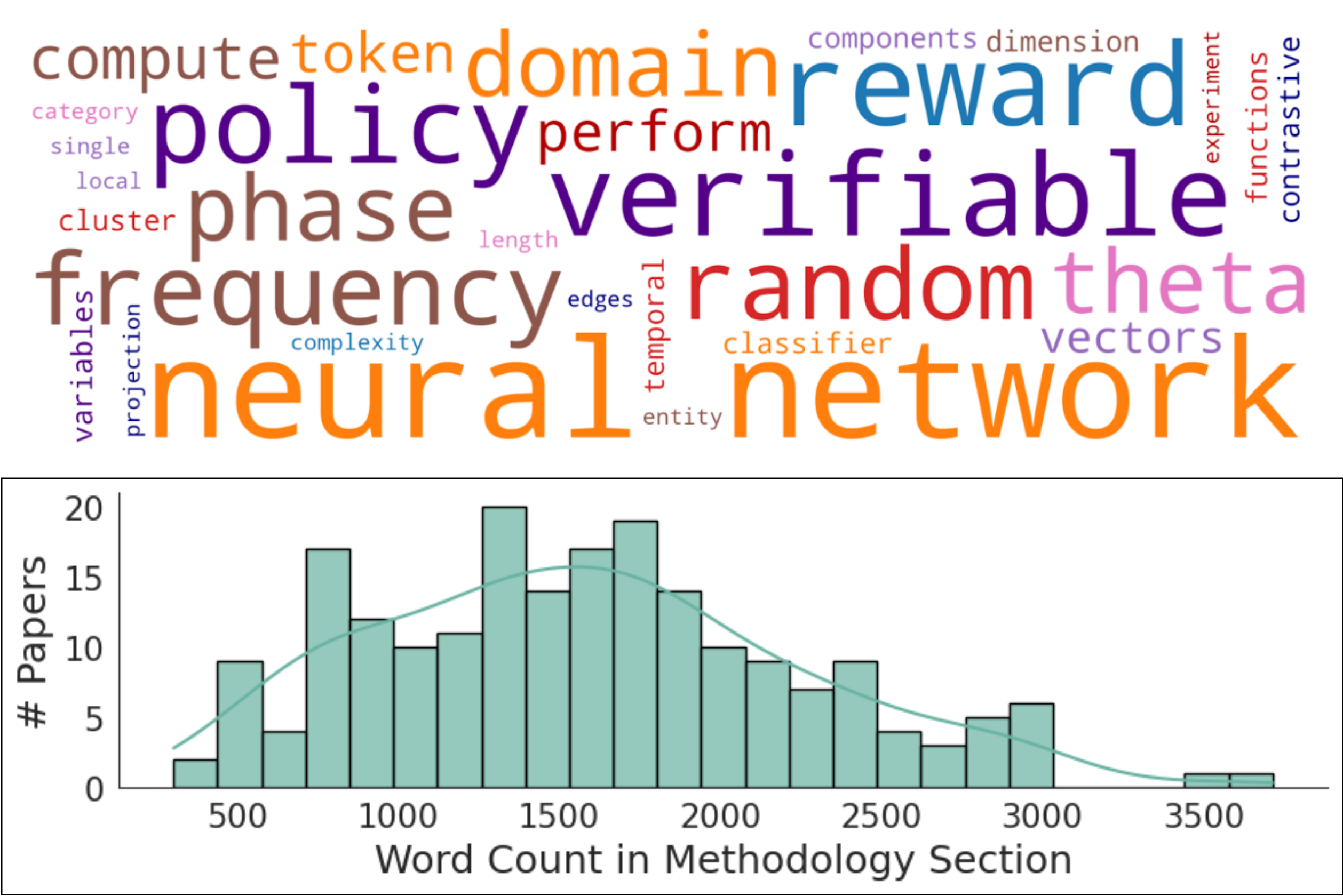}
\caption{\textbf{Methodology section statistics of \data{}.} Top:
word cloud of frequent technical terms across the 435 methodology
sections. Bottom: histogram of word counts (415 -- 3673 words,
median 1219).}
\label{fig:word_stat}
\end{minipage}
\vspace{-5mm}
\end{figure*}

\vspace{1mm}
\noindent \textbf{Paper Collection.} We collect papers from 15 top-tier computer science conferences in 2023, spanning 37 arXiv domains (Figure~\ref{fig:paper_domains}). We retain papers with TeX sources downloadable via the Python arXiv API\footnote{Python arXiv API: https://github.com/lukasschwab/arxiv.py} and at least one figure, yielding 5,570 peer-reviewed papers with TeX sources, author-provided figures, and main-body text.

\vspace{1mm}
\noindent \textbf{Filtering.} We use a three-stage pipeline to extract methodology sections and figures. ($i$) \emph{Section and figure identification:} We prompt GPT-5.2~\cite{gpt5.2} over paper sections to identify the proposed-system description, discard papers without a methodology section, and manually verify cases with multiple candidates. We similarly identify the methodology figure from captions and verify that it is cited in the methodology section. ($ii$) \emph{Basic quality check:} We remove likely artifacts: extreme-aspect-ratio figures (height $>$ 4$\times$width or width $>$ 5$\times$height) and images below $200\times200$ pixels. ($iii$) \emph{Text--figure consistency:} We ask Gemini-3.1-Pro~\cite{gemini3.1pro} and GPT-5.2 whether the figure accurately depicts the extracted method text, discarding samples rejected by either model to prioritize precision. This yields 2,632 papers; all prompts are in the appendix \ref{appendix:filter_prompt}.

\vspace{1mm}
\noindent \textbf{Human Verification.} We verify the dataset with 60 AI experts with active publication records. Each figure--text--caption sample is reviewed by two annotators for visual quality, readability, structural completeness, and text alignment, with a binary keep/remove decision; disagreements are resolved by a third annotator. This yields 2,219 valid samples, from which we randomly select 435 for the final evaluation set and use the remaining 1,784 to develop evaluation rubrics (Section~\ref{sec:evaluation_protocol}). We show domain distribution in Figure~\ref{fig:paper_domains}, methodology-section statistics in Figure~\ref{fig:word_stat}, and additional curation and annotator details in the appendix \ref{appendix:verification_annotators}.

\vspace{-2mm}
\subsection{\eval{}: Evaluation Protocol} \label{sec:evaluation_protocol}
\vspace{-2mm}

Evaluating generated methodology figures is challenging because quality is multi-dimensional: a figure should be correct, complete, readable, accessible, visually coherent, and structurally aligned with the described method. The same method may admit multiple valid layouts, making reference comparison useful but incomplete, while single-VLM judging can introduce bias. We therefore propose \eval{}, which evaluates generated figures along four complementary components:

\vspace{1mm}
\noindent \textbf{Completeness and Correctness.} A methodology figure should convey the system's key components and information flow. Inspired by PaperQuiz~\cite{paper2poster}, we give the methodology text and author-drawn figure to Claude-Opus-4.6~\cite{claude4.6}, which generates 50 yes/no questions with ground-truth answers across 10 conceptual dimensions. Gemini-3.1-Pro~\cite{gemini3.1pro}, GPT-5.2~\cite{gpt5.2}, and Qwen3-VL~\cite{bai2025qwen3vl} answer these questions using only generated figure; we report their average accuracy. Example questions are in appendix \ref{appendix:cc_questions}.

\begin{table*}[!t]
  \centering
  \small
  \setlength{\tabcolsep}{4pt}
  \resizebox{0.85\textwidth}{!}{
  \begin{tabular}{l | c | c c c c | c c c c c}
  \toprule
  \multirow{2}{*}{\bf Model} & \multirow{2}{*}{\bf C\&C} & \multicolumn{4}{c|}{\bf Reference-Based Fidelity} & \multicolumn{5}{c}{\bf Rubric-Based Content Quality} \\
  & & \bf Sim & \bf Rel & \bf OCR-F1 & \bf Avg & \bf Tech & \bf Clarity & \bf Cons & \bf Read & \bf Avg \\
  \midrule
  \multicolumn{11}{c}{\textit{Oracle Systems}} \\
  \midrule
  GT Figures & 6.4 & $-$ & 1.4 & $-$ & $-$ & 7.3 & 7.2 & \underline{8.0} & 6.8 & 7.3 \\
  \midrule
  \multicolumn{11}{c}{\textit{Single-Agent Systems}} \\
  \midrule
  GPT-5.4-Image~\cite{gpt5.2} & 3.7 & 5.9 & 1.3 & 1.5 & 2.9 & 5.8 & 6.8 & 6.4 & 5.5 & 6.1 \\
  Qwen-Image~\cite{wu2025qwenimage} & 2.2 & 4.3 & \textbf{1.6} & 0.3 & 2.1 & 3.9 & 3.7 & 4.2 & 3.6 & 3.9 \\
  Nano Banana Pro~\cite{nanobananapro} & 7.0 & 6.9 & \underline{1.5} & 1.7 & 3.4 & \textbf{7.7} & \underline{7.3} & 7.8 & \textbf{7.1} & \underline{7.5} \\
  \midrule
  \multicolumn{11}{c}{\textit{Agentic Frameworks}} \\
  \midrule
  Paper2Poster~\cite{paper2poster} & 2.3 & 2.3 & \underline{1.5} & 1.2 & 1.7 & 3.3 & 5.0 & 2.8 & 3.0 & 3.5 \\
  AutoFigure~\cite{zhu2026autofigure} & 6.3 & 5.6 & 1.1 & 1.6 & 2.8 & 6.7 & 6.6 & 6.9 & 6.3 & 6.6 \\
  Paper2SysArch~\cite{guo2025paper2sysarch} & 4.2 & 4.3 & 1.4 & 1.9 & 2.5 & 5.8 & 6.6 & 6.2 & 5.2 & 6.0 \\
  PaperBanana~\cite{zhu2026paperbanana} & \underline{7.1} & \underline{7.2} & 1.3 & \underline{2.4} & \underline{3.7} & \underline{7.4} & 7.2 & 7.5 & \underline{7.0} & 7.3 \\
  \rowcolor{Light}
  \method{} & \textbf{7.5} & \textbf{7.8} & \textbf{1.6} & \textbf{2.6} & \textbf{4.0} & \underline{7.4} & \textbf{7.4} & \textbf{8.6} & \underline{7.0} & \textbf{7.7} \\
  \midrule
  \textbf{\textcolor{blue}{$\Delta_{\text{\method\ - PaperBanana}}$}} & \textcolor{blue}{$0.4\uparrow$} & \textcolor{blue}{$0.6\uparrow$} & \textcolor{blue}{$0.3\uparrow$} & \textcolor{blue}{$0.2\uparrow$} & \textcolor{blue}{$0.3\uparrow$}
  & \textcolor{blue}{$0.0$} & \textcolor{blue}{$0.2\uparrow$} & \textcolor{blue}{$1.1\uparrow$} & \textcolor{blue}{$0.0$} & \textcolor{blue}{$0.4\uparrow$} \\
  \bottomrule
  \end{tabular}
  }
  \vspace{-1mm}
  \caption{\textbf{Main results on \data{} across three \eval{} axes (0--10 scale).} \textbf{C\&C} measures information coverage with cross-model VQA over 50 yes/no questions. \textbf{Reference-Based Fidelity} uses Sim (DINOv2), Rel (SigLIP2), and OCR-F1 against the author-drawn reference. \textbf{Rubric-Based Content Quality} averages four binary rubrics from 1{,}784 real figures: Tech, Clarity, Cons, and Read. See Section~\ref{sec:evaluation_protocol} for definitions.}
  \label{tab:main_results_rubric}
  \vspace{-3mm}
  \end{table*}

\vspace{1mm}
\noindent \textbf{Reference-Based Fidelity.} We compare each generated figure with the author-drawn reference using three metrics: ($i$) \emph{Visual similarity}, DINOv2-Large~\cite{oquab2024dinov2} \texttt{CLS} cosine similarity, which captures diagram structure without the pixel-level brittleness of SSIM~\cite{wang2004ssim}; we use DINOv2 rather than CLIP~\cite{radford2021clip} because its self-supervised features better capture spatial layout and local relationships. ($ii$) \emph{Figure-text relevance}, SigLIP2-Large~\cite{tschannen2025siglip2} cosine similarity between the generated figure and methodology text. ($iii$) \emph{OCR text F1}, token-level F1 between labels extracted from both figures using GOT-OCR~2.0~\cite{wei2024gotocr}. These metrics assess visual, cross-modal, and textual fidelity.

\vspace{1mm}
\noindent \textbf{Rubric-Based Content Quality.} We use VLM-as-a-judge to assess technical accuracy, visual clarity, structural consistency, and readability. Prior poster~\cite{paper2poster, zhang2025postergen}, figure~\cite{guo2025paper2sysarch, zhu2026paperbanana, zhu2026autofigure}, slide~\cite{zeng2025slidetailor, liang2025slidegen, zheng2025pptagent, ge2025autopresent}, and webpage~\cite{chen2025paper2web} work prompts a VLM for scale-based scores, making results sensitive to judge choice and scale range. Instead, we derive explicit rubrics from 1,784 held-out high-quality methodology figures (Section~\ref{sec:data_curation}). Each of the four aspects contains five yes/no questions, for 20 total, answered by Claude-Opus-4.6 given the generated figure and methodology text. We average answers per aspect and report the mean as a normalized 0--1 score; binary questions reduce scale sensitivity, while figure-derived rubrics align criteria with human-drawn diagrams. Details and examples are in the appendix \ref{appendix:rubric_questions}.

\vspace{1mm}
\noindent \textbf{Perceptual Design Quality.} Since VLMs are unreliable for low-level properties such as color contrast and pixel-level alignment, we add 8 deterministic heuristics grounded in information visualization~\cite{tufte1983visualdisplay, ware2019information} and graphic design: ($i$) \emph{Color \& Contrast}, using WCAG~2.1 accessibility~\cite{fernandez2019web21} and Matsuda-wheel hue harmony~\cite{matsuda1995color}; ($ii$) \emph{Density \& Clutter}, using canvas utilization and Tufte-style data-ink ratio~\cite{tufte1983visualdisplay}; ($iii$) \emph{Layout \& Alignment}, using pairwise non-overlap and whitespace uniformity; and ($iv$) \emph{Typography}, using font-size and font-style consistency. These metrics rely on pixel- and component-level statistics without a reference figure or judge VLM, making them fast and model-invariant. Further details are in the appendix \ref{appendix:perceptual_heuristics}.

\vspace{-2mm}
\section{Experiments}
\vspace{-2mm}

\begin{table*}[!t]
\centering
\small
\setlength{\tabcolsep}{6pt}
\resizebox{0.88\textwidth}{!}{
\begin{tabular}{l | c c | c c | c c | c c | c}
\toprule
\multirow{3}{*}{\bf Model} & \multicolumn{9}{c}{\bf Perceptual Design Quality} \\
& \multicolumn{2}{c|}{\bf Color \& Contrast} & \multicolumn{2}{c|}{\bf Density \& Clutter} & \multicolumn{2}{c|}{\bf Layout \& Align.} & \multicolumn{2}{c|}{\bf Typography} & \\
& \bf Access & \bf Hue & \bf Canvas & \bf Ink & \bf Overlap & \bf Uniform & \bf Font & \bf Text & \bf Avg \\
\midrule
\multicolumn{10}{c}{\textit{Oracle Systems}} \\
\midrule
GT Figures & 9.8 & 9.7 & 9.6 & 9.3 & 9.7 & 8.0 & 8.1 & 8.8 & 9.1 \\
\midrule
\multicolumn{10}{c}{\textit{Single-Agent Systems}} \\
\midrule
GPT-5.4-Image~\cite{gpt5.2} & 9.4 & 9.9 & 7.5 & 6.2 & 9.8 & 7.3 & 8.8 & 8.2 & 8.4 \\
Qwen-Image~\cite{wu2025qwenimage} & 4.3 & 9.9 & 9.6 & 2.3 & 9.7 & 4.2 & 7.9 & 8.1 & 7.0 \\
Nano Banana Pro~\cite{nanobananapro} & 9.1 & 9.8 & 9.6 & 3.8 & 9.9 & 6.2 & 8.4 & 8.2 & 8.1 \\
\midrule
\multicolumn{10}{c}{\textit{Agentic Frameworks}} \\
\midrule
Paper2Poster~\cite{paper2poster} & \textbf{10.0} & \textbf{10.0} & \textbf{10.0} & 7.6 & \textbf{10.0} & \textbf{8.8} & 8.6 & 8.4 & \underline{9.2} \\
AutoFigure~\cite{zhu2026autofigure} & 8.4 & 9.7 & 8.6 & 5.1 & 9.9 & 6.0 & 7.9 & 8.1 & 8.0 \\
Paper2SysArch~\cite{guo2025paper2sysarch} & 9.7 & 9.3 & 9.7 & \underline{7.9} & \textbf{10.0} & 4.6 & \textbf{10.0} & \underline{9.2} & 8.8 \\
PaperBanana~\cite{zhu2026paperbanana} & 9.1 & 9.8 & 9.4 & 4.6 & 9.9 & 6.5 & 8.5 & 8.1 & 8.2 \\
\rowcolor{Light}
\method{} & \underline{9.9} & \underline{9.9} & \underline{9.9} & \textbf{9.5} & \textbf{10.0} & \underline{8.4} & \textbf{10.0} & \textbf{9.8} & \textbf{9.7} \\
\midrule
\bf \textcolor{blue}{$\Delta_{\text{\method{} - PaperBanana}}$} & \textcolor{blue}{$0.8\uparrow$} & \textcolor{blue}{$0.1\uparrow$} & \textcolor{blue}{$0.5\uparrow$} & \textcolor{blue}{$4.9\uparrow$} & \textcolor{blue}{$0.1\uparrow$} & \textcolor{blue}{$1.9\uparrow$} & \textcolor{blue}{$1.5\uparrow$} & \textcolor{blue}{$1.7\uparrow$} & \textcolor{blue}{$1.5\uparrow$} \\
\bottomrule
\end{tabular}
}
\vspace{-1mm}
\caption{\textbf{Perceptual design quality on \data{} (0--10 scale).}
Eight deterministic heuristics in four axes: \textbf{Color \& Contrast} (Access: WCAG contrast; Hue: Matsuda harmony), \textbf{Density \& Clutter} (Canvas: canvas utilization; Ink: Tufte data-ink ratio), \textbf{Layout \& Alignment} (Overlap: pairwise non-overlap; Uniform: whitespace balance), and
\textbf{Typography} (Font: size consistency; Text: style consistency).
Full definitions in Section~\ref{sec:evaluation_protocol}.}
\label{tab:main_results_heuristics}
\vspace{-3mm}
\end{table*}

\begin{table}[t]                                                                                                         
\centering
\small                                                                                                                   
\setlength{\tabcolsep}{4pt}                                 
\resizebox{0.47\textwidth}{!}{\begin{tabular}{l | c | c c c c c}
\toprule                                                                                                                 
\textbf{Configuration} & \textbf{C\&C} & \textbf{Tech} & \textbf{Clarity} & \textbf{Cons} & \textbf{Read} & \textbf{Avg} \\
\midrule                                                                                                                 
Baseline & 5.2 & 5.5 & 5.3 & 6.0 & 5.2 & 5.5 \\             
+ Layout Agent & 5.8 & 6.0 & 5.8 & 6.5 & 5.7 & 6.0 \\
+ Feedback Agent & 6.2 & 6.5 & 6.3 & 7.1 & 6.1 & 6.5 \\
\midrule                                                                                                                 
+ CoT Round 1 & 6.6 & 6.8 & 6.7 & 7.6 & 6.5 & 6.9 \\
+ CoT Round 2 & 6.9 & 7.1 & 7.0 & 8.0 & 6.8 & 7.2 \\
+ CoT Round 3 & \underline{7.3} & \underline{7.3} & \underline{7.3} & \underline{8.4} & \underline{6.9} & \underline{7.5} \\
\rowcolor{Light}
+ CoT Round 4 & \textbf{7.5} & \textbf{7.4} & \textbf{7.4} & \textbf{8.6} & \textbf{7.0} & \textbf{7.7} \\
\bottomrule                                                                                                              
\end{tabular}}
\vspace{-1mm}
\caption{\textbf{Ablation study of \method{}.} We evaluate each component's contribution on \textbf{C\&C} and the four \textbf{Rubric-Based Content Quality} rubrics (Tech, Clarity, Cons, Read), starting from only the description and component agents and adding the layout agent, feedback agent, and CoT refinement rounds. The final row (CoT Round~4) matches our main configuration in Table \ref{tab:main_results_rubric}.}
\vspace{-6mm}
\label{tab:ablation}
\end{table}

\subsection{Baselines}

We compare \method{} with seven baselines: three single-agent raster generators---GPT-5.4-Image~\cite{gpt5.2}, Nano Banana Pro~\cite{nanobananapro}, and Qwen-Image~\cite{wu2025qwenimage}---and four agentic pipelines: Paper2Poster~\cite{paper2poster}, AutoFigure~\cite{zhu2026autofigure}, PaperBanana~\cite{zhu2026paperbanana}, and Paper2SysArch~\cite{guo2025paper2sysarch}. For Paper2Poster, we adapt only the prompts to generate methodology figures instead of posters. All baselines use the same methodology text and figure description.

\subsection{Implementation Details}

\method{} uses Claude-Sonnet-4.6~\cite{claude4.6} for planning and layout, Nano Banana Pro~\cite{nanobananapro} for PNG-mode components, draw.io XML for textbox-mode elements, and Gemini-3.1-Pro~\cite{gemini3.1pro} for $K=4$ feedback rounds. For \eval{}, Claude-Opus-4.6~\cite{claude4.6} generates questions and scores rubrics, while Gemini-3.1-Pro~\cite{gemini3.1pro}, GPT-5.2~\cite{gpt5.2}, and Qwen3-VL~\cite{bai2025qwen3vl} answer questions for completeness and correctness. Generation takes about 10 minutes per paper and costs approximately \$1.2 including full \eval{} (\$525 for 435 papers). Prompts, hyperparameters, and additional implementation details are provided in the appendix \ref{appendix:impl}.

\subsection{Results}

Tables~\ref{tab:main_results_rubric} and~\ref{tab:main_results_heuristics} report \method{} and baseline performance across the four \eval{} axes; the \textbf{best} score in each column is bolded and the \underline{second-best} is underlined.

\vspace{1mm}
\noindent \textbf{Completeness and Correctness.} \method{} achieves the highest C\&C score, improving over PaperBanana by $0.4$ points and Nano Banana Pro by $0.5$ points. This suggests that \method{} produces figures whose components and information flow are easier for VLMs to recover from the visual output. The author-drawn figures score $6.4$ under this protocol, likely because real paper figures often do not show all important fine-grained aspects of the proposed system for simplicity. While Nano Banana Pro and PaperBanana remain competitive, GPT-5.4-Image, Qwen-Image, Paper2Poster, and Paper2SysArch score much lower, indicating more frequent omissions or distortions of method components.

\vspace{1mm}
\noindent \textbf{Reference-Based Fidelity.} \method{} achieves the best reference-fidelity average, reaching $4.0$ compared with $3.7$ for PaperBanana. It also leads on all three sub-metrics, with visual similarity of $7.8$, figure-text relevance of $1.6$, and OCR-F1 of $2.6$. These gains indicate that \method{} better preserves both the structure and textual content of author-drawn references. The OCR-F1 improvement is especially important: native draw.io text keeps labels, symbols, and equations editable and legible, whereas raster image generators often blur or corrupt scientific notation.


\begin{table}[t]
\centering
\small
\vspace{-2.5mm}
\begin{tabular}
{@{}p{0.54\columnwidth}@{\hspace{0.02\columnwidth}}p{0.42\columnwidth}@{}}

\begin{minipage}[t]{\linewidth}
\vspace{0pt}
\centering
\setlength{\tabcolsep}{3.2pt}
\renewcommand{\arraystretch}{0.92}
\resizebox{\linewidth}{!}{
\begin{tabular}{@{}c l r r@{}}
\toprule
\textbf{R} & \textbf{Method} & \textbf{Elo} & \textbf{Win\%} \\
\midrule
\cellcolor{Light}1 & \cellcolor{Light}\method{} & \cellcolor{Light}1327 & \cellcolor{Light}81.4 \\
2 & Nano Banana Pro & 1201 & 74.6 \\
3 & PaperBanana     & 1175 & 71.3 \\
4 & AutoFigure      & 1058 & 56.1 \\
5 & GT Figures      & 1044 & 54.4 \\
6 & GPT-5.4-Image   &  909 & 36.6 \\
7 & Paper2SysArch   &  879 & 33.2 \\
8 & Paper2Poster    &  577 &  7.3 \\
\bottomrule
\end{tabular}
}
\end{minipage}

&

\begin{minipage}[t]{\linewidth}
\vspace{0pt}
\caption{\textbf{Human evaluation.} Elo ratings and win rates from 10{,}000 pairwise comparisons over 435 papers. \method{} ranks first overall.}
\label{tab:human_eval}
\end{minipage}

\end{tabular}
\vspace{-10.5mm}
\end{table}

\begin{figure*}[!t]
\centering
\includegraphics[width=0.999\textwidth]{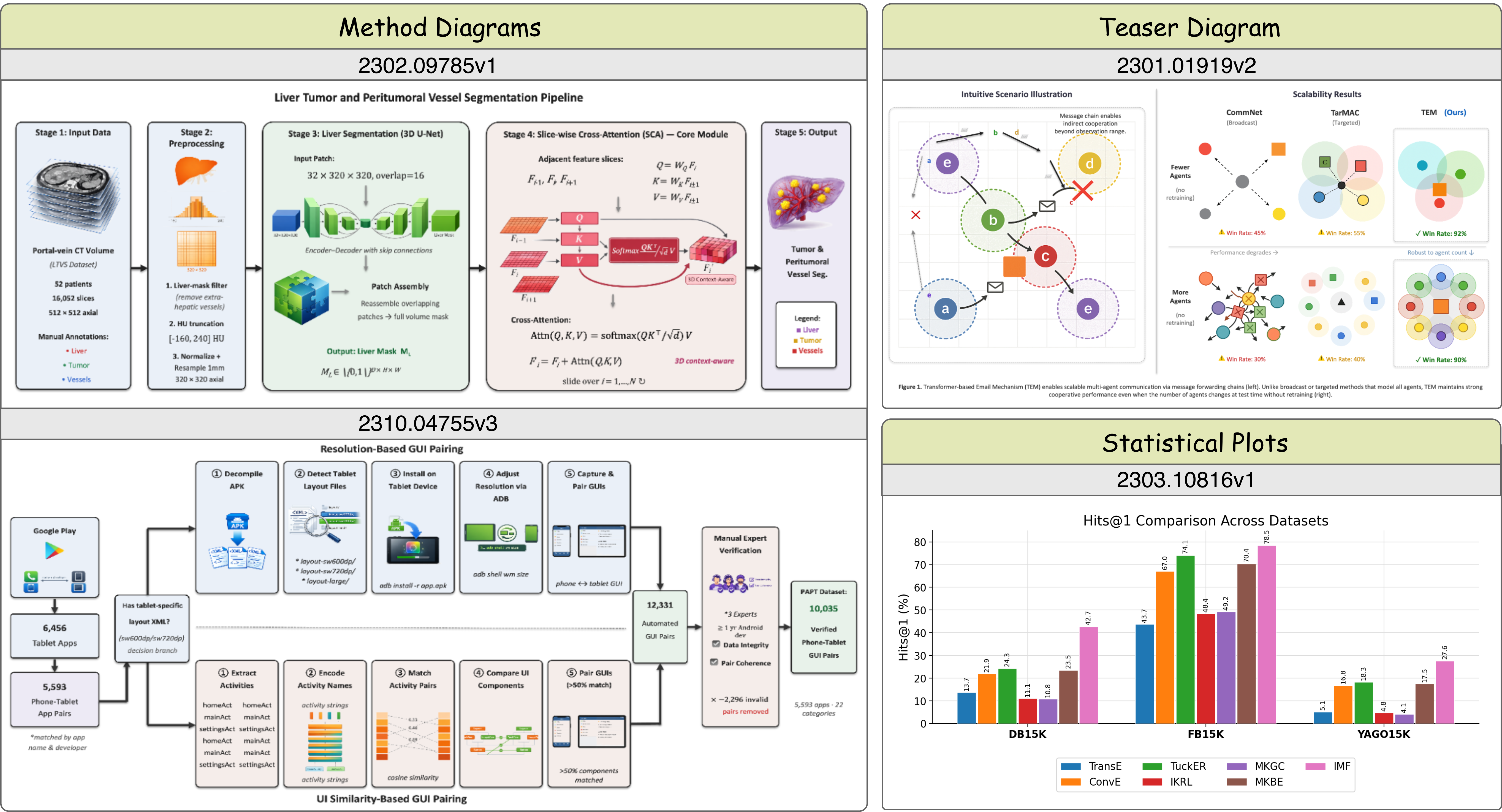}
\vspace{-6mm}
\caption{\textbf{Qualitative examples from \method{}.} Method diagrams on the left show hierarchical multi-stage pipelines; the teaser diagram and statistical plot on the right illustrate generalization beyond methodology figures.}
\vspace{-3mm}
\label{fig:qual}
\end{figure*}

\vspace{1mm}
\noindent \textbf{Rubric-Based Content Quality.} \method{} obtains the highest rubric-based average, scoring $7.7$ compared with $7.5$ for Nano Banana Pro and $7.3$ for author-drawn figures. Its strongest gains appear in clarity and structural consistency, where it reaches $7.4$ and $8.6$, respectively. The consistency score also exceeds the author-drawn figures, suggesting that explicit planning and structured layout generation help organize method components more uniformly. Nano Banana Pro remains competitive in technical accuracy and readability, but its lower consistency score of $7.8$ shows that visual polish alone does not always yield a coherent representation of modular information flow.

\vspace{1mm}
\noindent \textbf{Perceptual Design Quality.} In Table~\ref{tab:main_results_heuristics}, \method{} achieves the highest perceptual-design average of $9.7$, ahead of Paper2Poster at $9.2$ and Paper2SysArch at $8.8$. It ranks among the top two methods on every heuristic and leads on ink usage, overlap, font consistency, and text contrast. The large $4.9$ gain in ink usage over PaperBanana highlights the advantage of editable vector layouts, which allocate canvas space more efficiently than rasterized image-generation outputs. Overall, these results show that \method{} improves both semantic fidelity and low-level design properties such as spacing, typography, and layout cleanliness.

\vspace{-2mm}
\subsection{Ablation Study}
\vspace{-1mm}

Table~\ref{tab:ablation} quantifies the contribution of each component in \method. Starting from description and component agents, adding the layout agent improves average score from $5.5$ to $6.0$, showing the value of explicit layout reasoning. Adding the feedback agent further raises the score to $6.5$, validating the iterative refinement.

The bottom block varies the number of CoT refinement rounds $K$. Performance improves from $6.9$ at $K=1$ to $7.7$ at $K=4$, with gains tapering after the third round; we therefore use $K=4$ in the main configuration. Structural consistency benefits most, rising from $6.0$ in the baseline to $8.6$, indicating that refinement effectively resolves layout-level errors left by single-pass generation.

\vspace{-2mm}
\subsection{Human Evaluation}
\vspace{-1mm}

We conduct a blind pairwise human study with 60 participants with median research experience of 2 years. In each trial, a participant sees two anonymized figures for the same paper, generated by two randomly selected methods, and selects the better one based on technical accuracy, visual clarity, structural consistency, and readability. From 10{,}000 pairwise comparisons over 435 papers, we compute Elo rankings. As shown in Table~\ref{tab:human_eval}, \method{} ranks first with an Elo rating of 1327 and an 81.4\% win rate, outperforming the author-drawn GT figures and all evaluated baselines while preserving full editability. \eval{}'s rubric-based content quality strongly aligns with human judgments, with Pearson correlation $r{=}0.92$ ($p{<}0.001$) for structural consistency and $r{=}0.83$ ($p{<}0.01$) for technical accuracy.

\vspace{-2mm}
\subsection{Visualization \& Error Analysis}
\vspace{-1mm}

Figure~\ref{fig:qual} presents qualitative examples generated by \method{}, spanning our primary setting and broader scientific visualizations. The two examples on the left show dense methodology diagrams with multi-stage pipelines, hierarchical modules, heterogeneous visual components, legends, and labeled information flow, while the right examples extend to teaser-style diagrams and statistical plots. Although our benchmark and evaluation focus on methodology figures, these examples suggest broader generalization. Remaining errors are mostly local, such as minor overlaps and misalignments, and could be reduced with finer geometric constraints; we provide examples of failure cases in appendix \ref{appendix:failure_cases}.

\vspace{-2mm}
\section{Conclusion}
\vspace{-2mm}

We present \method{}, a multi-agent framework for generating visually rich and fully editable methodology figures, together with \data{}, a human-verified benchmark, and \eval{}, a four-axis evaluation protocol. Across seven single- and multi-agent baselines, \method{} achieves the strongest performance on all \eval{} axes and ranks first in blind pairwise human evaluation, while generating editable figures under 10 minutes on average. Qualitative examples further suggest that the framework can extend beyond methodology figures to teaser diagrams and statistical plots.

\section*{Limitations}

We discuss several limitations of our work.

\vspace{1mm}
\noindent \textbf{Scope of figure types and domains.} \method{} is designed primarily for methodology figures, a central and challenging form of scientific visualization. Although the qualitative examples in Figure~\ref{fig:qual} show generalization to teaser diagrams and statistical plots, we do not systematically benchmark \method{} on these figure types. In addition, \data{} is drawn exclusively from AI/ML venues, covering areas such as computer vision, natural language processing, machine learning, KDD/data mining, and signal processing. It does not include papers from materials science, chemistry, physics, biology, or other scientific communities, where visual conventions can differ substantially, such as chemical structures, anatomical diagrams, particle-physics diagrams, and microscopy panels. Supporting these domains may require additional component libraries or domain-specific adaptation.

\vspace{1mm}
\noindent \textbf{Model dependence.} Although \eval{} reduces reliance on any single judge through cross-model VQA, deterministic heuristics, and reference-based metrics, several axes still depend on learned model outputs and may shift as the underlying models evolve. A fully model-independent evaluation of methodology-figure quality remains an open problem.

\vspace{1mm}
\noindent \textbf{Cost and compute.} Each figure costs approximately \$1.2 in API calls and takes about 10 minutes to generate with $K{=}4$ refinement rounds, which may be prohibitive at very large scale. Distilled or open-weight backbones could reduce cost, but may also reduce output quality.

\vspace{1mm}
\noindent \textbf{Remaining failure modes.} Despite iterative refinement, some figures still exhibit minor component overlap, misalignment, or unclear arrow routing. We document representative failure cases in the appendix \ref{appendix:failure_cases}; tighter geometric constraints or improved spatial reasoning could further reduce these errors.

\vspace{1mm}
\noindent \textbf{Editability boundary.} While \method{} outputs can be opened in standard diagramming tools, substantial revisions still require some familiarity with those tools. Natural-language editing interfaces may further lower this barrier in future work.

\vspace{1mm}
\noindent \textbf{Model versioning.} \method{} relies on proprietary LLMs and VLMs, including Claude, Gemini, Nano Banana Pro, and GPT, whose behavior may change across versions. We document the exact model versions and input prompts in the appendix \ref{appendix:impl}, \ref{appendix:cc_generation_prompt} and \ref{appendix:filter_prompt}, but full reproducibility may require access to the same model snapshots.

\section*{Ethical Considerations}

\noindent \textbf{Reproducibility.} We provide detailed agent prompts, hyperparameters, the full \eval{} protocol with questions and rubrics, per-component cost breakdowns, and \data{} in the appendix \ref{appendix:impl} to facilitate reproduction. The source code of \method{} and \eval{} will be released for research use upon acceptance.

\vspace{1mm}
\noindent \textbf{Data sources.} All papers in \data{} are drawn from arXiv preprints, which are distributed under arXiv-hosted licenses, including the arXiv non-exclusive distribution license and Creative Commons variants such as CC~BY~4.0, CC~BY-SA~4.0, and CC~BY-NC-SA~4.0. We release benchmark metadata, including arXiv IDs, captions, and extracted methodology text, rather than the original PDF or figure files. This avoids indefinitely redistributing authors' original figures if a preprint is later removed.

\vspace{1mm}
\noindent \textbf{Annotation.} Our 60 annotators include undergraduates, master's students, PhD students, and one postdoctoral researcher with AI/ML research experience. Annotators provided informed consent and were compensated at or above the standard local rate for academic research assistance. We conducted clarification sessions before each annotation round and reproduce all annotation instructions verbatim in the appendix \ref{appendix:filter_prompt} and \ref{appendix:verification_annotators}.

\vspace{1mm}
\noindent \textbf{Biases.} \data{} is drawn from English-language AI/ML venues from 2023 and therefore inherits the stylistic norms of this community, such as neural-network blocks and attention diagrams. Generated figures may not generalize equally well to other scientific communities or languages. We encourage future work to extend the benchmark beyond AI/ML and recommend interpreting quantitative results within this scope.

\vspace{1mm}
\noindent \textbf{Misuse potential.} Generated figures could be used to misrepresent methods, polish unsupported claims, or fabricate visual support for results. Authors using \method{} should verify that generated figures accurately reflect their methods, and reviewers should not treat figure polish as a proxy for technical correctness. \method{} is intended as an assistive drafting tool, not a substitute for authorial judgment.

\vspace{1mm}
\noindent \textbf{Intended use.} We release \method{}, \data{}, and \eval{} for research use only. Our goal is to support researchers' productivity, particularly for those without specialized design training, while preserving author control over the final figure.

\vspace{1mm}
\noindent \textbf{Environmental impact.} Each figure generation requires multiple API calls to large LLMs and VLMs, contributing to compute-related emissions. \method{} does not train models from scratch; all components use inference-only API calls, which limits its training-time environmental footprint compared with systems that require custom model training.

\bibliography{custom}

\newpage
\thispagestyle{plain}
\makeatletter
\twocolumn[\LARGE \bf \centering Appendix \par \bigskip
 ]
\appendix

\counterwithin{figure}{section}
\numberwithin{table}{section}
\label{sec:appendix}

\noindent \textbf{Appendix contents.} This appendix is organized as follows. Section~\ref{appendix:qual_results} provides additional qualitative results, including methodology figures, teaser diagrams, statistical plots, and per-round CoT refinement examples. Section~\ref{appendix:failure_cases} analyzes representative failure cases and localizes residual errors to component-level rendering. Section~\ref{appendix:impl} describes the four-agent implementation, agent prompts, and quality-control harness. Section~\ref{appendix:eval_questions} details the question sets and scoring protocol for the two VLM-answered \eval{} axes: Completeness and Correctness, and Rubric-Based Content Quality. Section~\ref{appendix:perceptual_heuristics} defines the eight deterministic Perceptual Design Quality heuristics, and Section~\ref{appendix:data_curation} reports the text--figure consistency filtering prompt and human-study details.







\section{Additional Qualitative Results} \label{appendix:qual_results}

\begin{figure*}[!t]
\centering
\includegraphics[width=\textwidth]{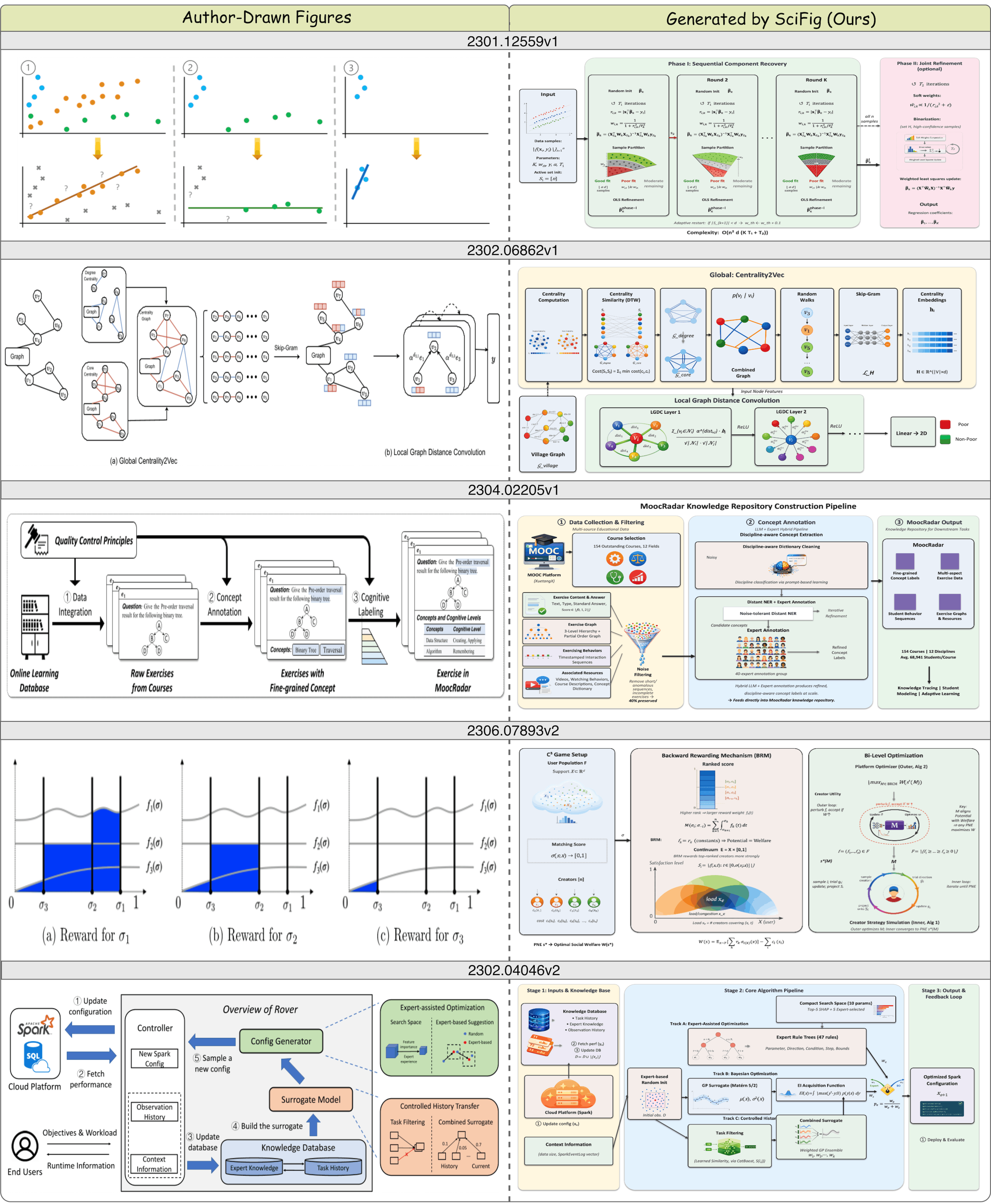}
\vspace{-2mm}
\caption{\textbf{Author-drawn methodology figures vs.\ \method{}-generated figures.} Each row shows the original author-drawn figure (left) and the \method{} output (right) for the same paper. \method{} makes intermediate components, sub-modules, variables, and inter-stage information flow explicit while preserving editability.}
\vspace{-3mm}
\label{fig:qual_main}
\end{figure*}

\begin{figure*}[!t]
\centering
\includegraphics[width=\textwidth]{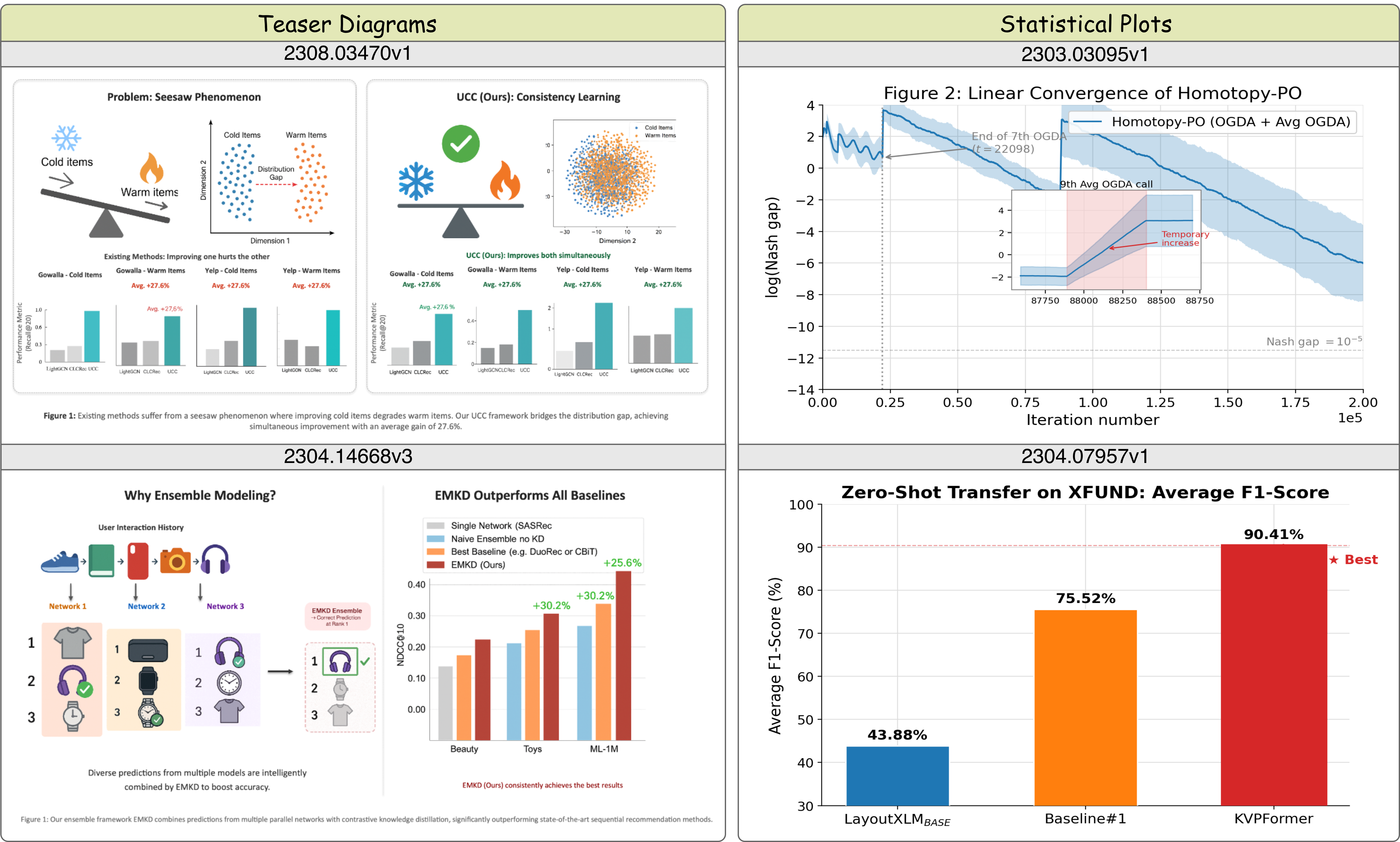}
\vspace{-5mm}
\caption{\textbf{\method{} generalizes to teaser diagrams and statistical plots.} Top: teaser-style conceptual diagrams. Bottom: statistical plots for convergence analysis and zero-shot transfer comparison. The same multi-agent pipeline produces editable, well-labeled outputs across diverse figure types without retraining.}
\vspace{-3mm}
\label{fig:qual_teaser}
\end{figure*}

\begin{figure}[!t]
\centering
\includegraphics[width=\columnwidth]{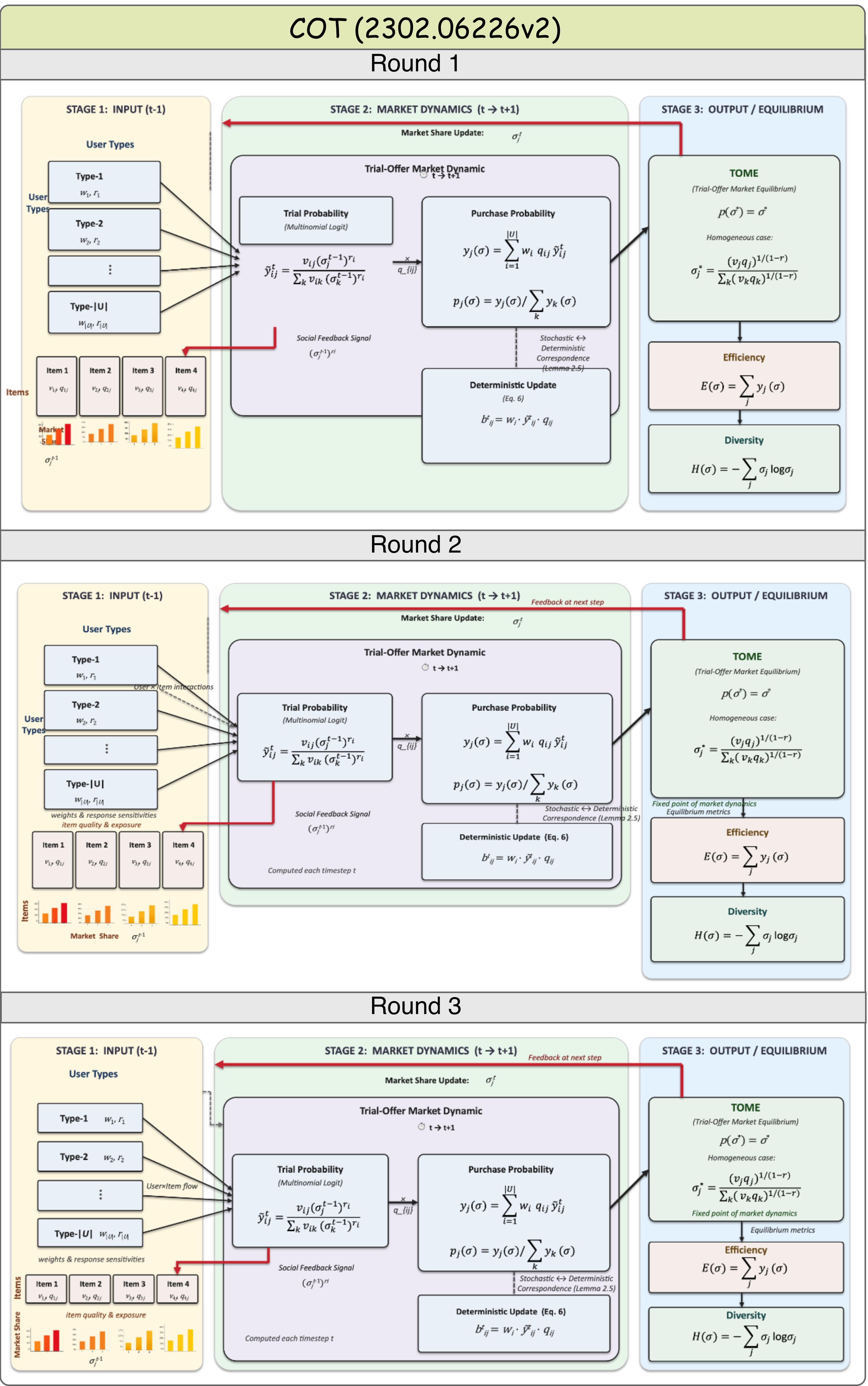}
\vspace{-5mm}
\caption{\textbf{Effect of iterative CoT refinement on a single figure.} For paper 2302.06226v2, Round~1 establishes the overall structure but contains misalignment and text overlap. Round~2 improves spacing and alignment, while Round~3 produces cleaner component placement, arrow routing, and fine-grained labels.}
\vspace{-4mm}
\label{fig:qual_cot}
\end{figure}

We provide additional qualitative examples highlighting \method{} across methodology diagrams and broader scientific visualizations. Figure~\ref{fig:qual_main} compares author-drawn figures with \method{} outputs for the same papers, Figure~\ref{fig:qual_teaser} shows teaser diagrams and statistical plots, and Figure~\ref{fig:qual_cot} illustrates the effect of iterative CoT refinement.

\vspace{1mm}
\noindent \textbf{Comparison with author-drawn figures.}
Figure~\ref{fig:qual_main} compares original author-drawn methodology figures (left) with \method{}-generated figures (right). Across these examples, \method{} makes intermediate components, sub-modules, variables, and inter-stage information flow more explicit. In contrast, author-drawn figures often compress multiple operations into single blocks or omit intermediate details for visual simplicity. Both styles can be useful, but these examples show that \method{} can produce detailed, editable diagrams that closely follow the underlying methodology text.

\vspace{1mm}
\noindent \textbf{Generalization to teaser diagrams and statistical plots.}
Figure~\ref{fig:qual_teaser} shows examples beyond methodology figures: two teaser-style conceptual diagrams on the top row and two statistical plots on the bottom row. These examples suggest that the same multi-agent pipeline can adapt to other scientific figure types without retraining, producing editable outputs with clear labels, appropriate visual elements, and structured layouts.

\vspace{1mm}
\noindent \textbf{Effect of CoT refinement rounds.}
Figure~\ref{fig:qual_cot} shows how iterative feedback improves a generated figure for paper 2302.06226v2. Round~1 captures the overall structure but contains stage misalignment and text-box overlap. Round~2 improves spacing and alignment, producing a more balanced layout. By Round~3, component placement, arrow routing, dense equations, and variable labels are more readable and visually organized. This progression illustrates how iterative feedback corrects layout-level issues that often remain in single-pass generation.

\section{Failure Cases and Error Analysis} \label{appendix:failure_cases}

\begin{figure}[!t]
\centering
\includegraphics[width=\columnwidth]{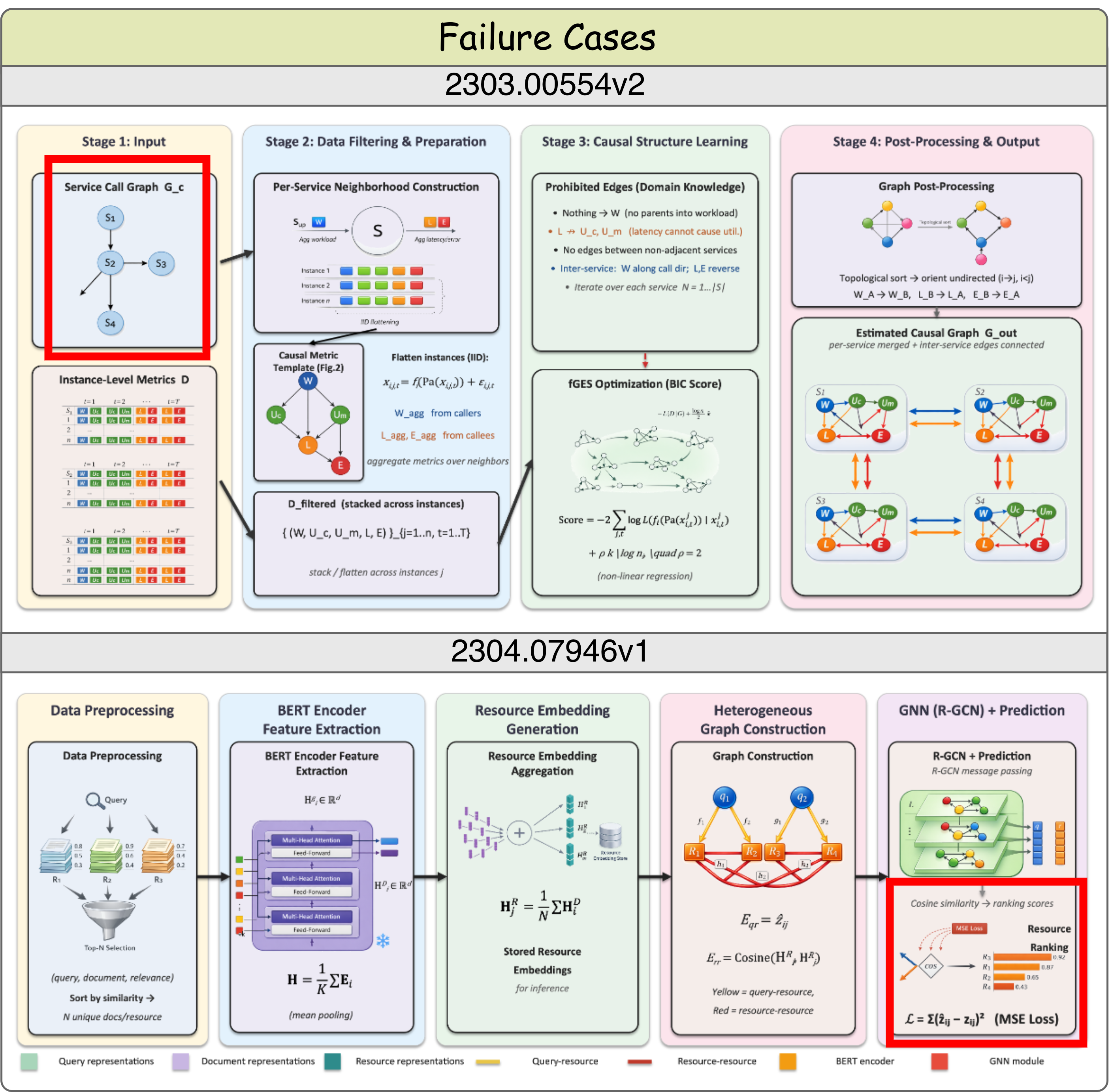}
\vspace{-5mm}
\caption{\textbf{Failure cases of \method{}.} The examples from papers 2303.00554v2 (top) and 2304.07964v1 (bottom) show errors localized within PNG-mode components: a missing graph node and an incorrectly rendered MRR-with-cosine-similarity formula. The surrounding layout and alignment remain correct, indicating that these failures originate from component rendering rather than layout refinement.}
\vspace{-4mm}
\label{fig:qual_failure}
\end{figure}

Although \method{} generally produces well-structured figures with coherent module organization and information flow, some errors remain. Figure~\ref{fig:qual_failure} shows two representative cases, from papers 2303.00554v2 and 2304.07964v1, where the overall layout, spacing, and arrow routing are visually reasonable, but individual component contents are incorrect. In the first case, the input graph diagram is missing a node, leading to an incomplete representation of the graph structure used by the method. In the second case, the MRR loss in the prediction module is rendered incorrectly with respect to cosine similarity, producing a notation-level error in the equation.

These failures arise from component-level generation rather than layout refinement. In PNG mode, each visual element is rendered from a local component description, so small semantic errors---such as missing nodes, incorrect formulas, or mislabeled variables---can persist across refinement rounds. The feedback process is effective at correcting spatial issues such as overlap, alignment, and routing, but it is less reliable at verifying the internal semantic correctness of rendered components. Addressing these errors may require a stronger component-level verification stage that checks each rendered element against its planned specification, or image-generation models with better symbolic and structural reasoning.

\section{Additional Implementation Details}
\label{appendix:impl}

\method{} is implemented as a four-agent pipeline coordinated by a deterministic Python harness. For each paper, the system takes the methodology text and figure caption as input, generates an editable diagram artifact, renders it for visual feedback, and iteratively refines the output. We save agent transcripts, intermediate renderings, component assets, and feedback reports for reproducibility.

\subsection{Model Configuration}
\label{appendix:models}

The planning and layout agents use Claude-Sonnet-4.6 with temperature $0.1$. The component agent renders visually rich PNG-mode elements using Nano Banana Pro, while textbox-mode elements are written directly into the editable diagram representation. The feedback agent uses Gemini-3.1-Pro to critique rendered figures. For \eval{}, Claude-Opus-4.6 generates questions and scores rubric items, while Gemini-3.1-Pro, GPT-5.2, and Qwen3-VL answer questions for the Completeness and Correctness axis. All benchmark runs use a fixed configuration, without per-paper prompt tuning or seed search.

\subsection{Agent Prompts}
\label{appendix:prompts}

\method{} relies on prompting rather than fine-tuning. Its prompts control four stages: component planning, layout generation, visual feedback, and iterative refinement. We include the most important prompt excerpts below; full prompts will be released with the code.

\paragraph{Component planning.}
The component-planning prompt identifies the visual elements needed for the figure and assigns each to either PNG mode, for visually rich components, or textbox mode, for simple labels and formula boxes.

\begin{promptbox}{Component planning prompt: selecting visual components and rendering modes}
You are planning the visual illustrations for a publication-quality pipeline diagram.

Given the pipeline description, list the key visual illustrations needed for the figure.

For each component, provide:
- a unique snake_case key,
- a concise visual description,
- whether it should be rendered as PNG or TEXTBOX.

Guidelines:
- Prefer one primary visual illustration per pipeline stage.
- Use PNG for neural network blocks, feature maps, rendered outputs, graphs, plots, images, or architecture diagrams.
- Use TEXTBOX only for simple labels, formula boxes, or legend items.
- Do not use TEXTBOX for elements that require visual structure.
\end{promptbox}

\paragraph{Layout generation.}
The layout prompt converts the plan into an editable diagram using geometry-aware helper functions for modules, components, arrows, text, equations, and embedded images.

\begin{promptbox}{Layout generation prompt: creating an editable methodology diagram}
You are creating a publication-quality methodology figure as an editable diagram, not a raster-only image.

Design principles:
- Use a clean scientific-figure style.
- Make large visual components the centerpiece of each module.
- Preserve the method's stages, components, and information flow.
- Use clear directional arrows and concise labels.
- Keep spacing, alignment, and typography consistent.

Canvas:
- Use a left-to-right pipeline layout unless the method suggests otherwise.
- Center the full diagram and avoid clutter.
- Do not place the paper caption inside the canvas.

Arrow hierarchy:
- Use thick arrows for main inter-stage flow.
- Use thinner arrows for local intra-module connections.
- Use dashed arrows for losses, supervision, or feedback paths.

Math:
- Render non-trivial equations with the math helper, not plain text boxes.
\end{promptbox}

\paragraph{Visual feedback.}
The feedback prompt asks a VLM to critique only visible problems in the rendered figure, reducing hallucinated edits.

\begin{promptbox}{Visual feedback prompt: identifying visible layout and readability issues}
You are reviewing a rendered methodology figure for visual quality and readability.

Rules:
- Report only issues that are clearly visible in the figure.
- Do not invent missing components or unsupported content.
- Prefer repositioning, resizing, rerouting, or relabeling over deletion.
- Preserve important labels, variables, equations, and annotations.

Check for:
- component overlap,
- text overlap or unreadable labels,
- poor spacing or misalignment,
- unclear arrow routing,
- weak visual hierarchy,
- cluttered regions,
- blurry or low-quality components.

Return concise, actionable feedback with the affected element and recommended fix.
\end{promptbox}

\paragraph{Iterative refinement.}
The refinement prompt converts feedback into targeted edits while preserving the existing structure whenever possible.

\begin{promptbox}{Iterative refinement prompt: applying targeted edits across rounds}
You are refining an editable scientific methodology figure.

Inputs:
1. the current diagram script,
2. the rendered figure,
3. visual feedback from the previous round.

Make targeted edits only. Do not rewrite the figure from scratch.

Priorities:
- Preserve the overall method structure and information flow.
- Improve readability, spacing, alignment, and arrow routing.
- Preserve labels, variables, equations, and callouts.
- Prefer small local edits over global restructuring.
- Avoid changes that introduce new overlaps or disrupt correct structure.

Output a complete revised editable diagram.
\end{promptbox}

\subsection{Quality-Control Harness}
\label{appendix:harness}

In addition to VLM feedback, \method{} uses a lightweight deterministic harness to detect common low-level rendering issues. The harness checks for text overlap, duplicated labels, unstable changes across refinement rounds, low-quality component renderings, and obvious alignment or spacing problems. It also routes specialized content to appropriate renderers: LaTeX equations are rendered with Matplotlib, graph-structured components with Graphviz, and illustration-style components with the image model. These checks complement VLM critiques and provide additional feedback to the refinement agent.

\subsection{Iterative Refinement}
\label{appendix:cot_settings}

The refinement loop runs for at most $K=4$ rounds. At each round, \method{} renders the current figure, obtains VLM feedback, applies harness-based checks, and asks the refinement agent to revise the editable diagram. The loop may terminate early when no actionable feedback remains and the figure passes the automatic checks. If the maximum number of rounds is reached, the system returns the best intermediate output based on the feedback and harness reports.

\section{Evaluation Question Sets}
\label{appendix:eval_questions}

\eval{} reports four axes in the main paper: \emph{Completeness and Correctness} (C\&C), \emph{Reference-Based Fidelity}, \emph{Rubric-Based Content Quality}, and \emph{Perceptual Design Quality} (Section~\ref{sec:evaluation_protocol}). Only the first and third axes use VLM-answered question banks. C\&C uses paper-specific questions to test whether a generated figure preserves the concrete content of the corresponding method, while Rubric-Based Content Quality uses a fixed set of rubric questions to assess general properties of high-quality methodology figures. The other two axes do not use question answering: Reference-Based Fidelity is computed from DINOv2, SigLIP2, and OCR-F1 comparisons against the author-drawn figure, and Perceptual Design Quality is computed from deterministic visual heuristics. This appendix documents the two question banks, provides representative examples, and describes the scoring protocol. The full per-paper C\&C question banks for all 435 benchmark papers, together with the 20 rubric questions, will be released with our code.

\begin{table*}[t]
\centering
\footnotesize
\setlength{\tabcolsep}{4pt}
\renewcommand{\arraystretch}{1.15}
\begin{tabularx}{0.98\textwidth}{@{}L{0.19\textwidth} C{0.10\textwidth} Y L{0.08\textwidth}@{}}
\toprule
\textbf{\eval{} axis} & \textbf{Question-based?} & \textbf{Scoring procedure} & \textbf{Details} \\
\midrule
Completeness \& Correctness
& Yes
& 50 binary yes/no questions per paper, generated by Claude-Opus-4.6 from the methodology text and author-drawn figure across 10 conceptual dimensions. Gemini-3.1-Pro, GPT-5.2, and Qwen3-VL answer using only the generated figure.
& App.~\ref{appendix:cc_questions} \\[3pt]

\midrule
Rubric-Based Content Quality
& Yes
& 20 binary yes/no questions total, with 5 questions for each aspect: Technical Accuracy, Visual Clarity, Structural Consistency, and Readability. The questions are derived from 1{,}784 held-out high-quality methodology figures and answered by Claude-Opus-4.6.
& App.~\ref{appendix:rubric_questions} \\[3pt]

\midrule
Reference-Based Fidelity
& No
& Computed using DINOv2 visual similarity, SigLIP2 figure--text relevance, and OCR-F1 against the author-drawn figure.
& Sec.~\ref{sec:evaluation_protocol} \\[3pt]
\midrule

Perceptual Design Quality
& No
& Computed using 8 deterministic heuristics covering color/contrast, density/clutter, layout/alignment, and typography.
& Sec.~\ref{sec:evaluation_protocol} \\
\bottomrule
\end{tabularx}
\caption{\textbf{Question-based and non-question-based components of \eval{}.} C\&C and Rubric-Based Content Quality use VLM-answered question banks, while Reference-Based Fidelity and Perceptual Design Quality use reference-based metrics and deterministic heuristics.}
\label{tab:eval_question_sets}
\end{table*}

All questions within a bank contribute equally to the corresponding axis score. We do not apply difficulty-based weighting, per-question weighting, or judge-specific weighting. We also separate question generation from answering whenever possible, so that no model is asked to answer questions it generated.

\subsection{Completeness and Correctness Questions}
\label{appendix:cc_questions}

The C\&C question bank evaluates whether a generated figure preserves the paper-specific content of the method. Each question names concrete entities from the corresponding methodology section, such as modules, variables, loss terms, transformations, stage ordering, or inter-module connections. This makes the metric sensitive to content-level errors that may not be captured by general design rubrics. For example, a visually clean figure that omits a critical module, mislabels a loss term, or routes information in the wrong direction will receive ``no'' answers on the relevant questions and therefore a lower C\&C score.

For each of the 435 benchmark papers, Claude-Opus-4.6 reads the full methodology text and author-drawn figure, then generates 50 yes/no questions with ground-truth answers. The questions are distributed across 10 conceptual dimensions: components present, named variables, ordering of stages, input/output correspondence, supervision signal, loss terms, module-level groupings, inter-module connections, dimensional transformations, and overall pipeline direction. The generation prompt requires each question to be grounded in the specific paper and answerable from the rendered figure alone, so the score reflects what the figure shows rather than what the paper text says.

At evaluation time, Gemini-3.1-Pro, GPT-5.2, and Qwen3-VL independently answer all 50 questions using only the generated figure. We compute each model's accuracy against the ground-truth answers and average the three accuracies to obtain the C\&C score. Below, we show representative questions from three benchmark papers spanning temporal-graph reduction, dynamic-market recommendation, and microservice causal-structure learning. All listed examples have ground-truth answer ``Yes'' for the corresponding author-drawn figure.

\begin{promptbox}{C\&C examples: temporal-graph reduction paper}
Paper 2301.10503v1 -- temporal-graph reduction. Each question is binary yes/no.

Q1 (Components). Does the diagram contain a single central vertex visually distinguished from the peripheral vertices?

Q2 (Named variables). Are the edge labels indexed by the temporal variables j and j' visible on the edges of the construction?

Q3 (Module-level grouping). Are the peripheral vertices arranged in a radial or star-like layout around the central vertex, as opposed to a flat row or grid?
\end{promptbox}

\begin{promptbox}{C\&C examples: dynamic-market recommendation paper}
Paper 2302.06226v2 -- dynamic-market recommendation. Each question is binary yes/no.

Q1 (Module-level grouping). Does the figure visually separate the theoretical optimization layer from the practical market-dynamics layer, with explicit connecting lines or aligned positioning between them?

Q2 (Named variables). Are the discrete time steps, such as t and t+1, labeled on the temporal progression of the diagram?

Q3 (Inter-module connections). Are the attention-allocation edges between user groups and items drawn with line thickness or intensity that varies to indicate preference strength?
\end{promptbox}

\begin{promptbox}{C\&C examples: microservice causal-structure learning paper}
Paper 2303.00554v2 -- microservice causal-structure learning. Each question is binary yes/no.

Q1 (Components present). Are the three named metric types -- load requests, resource utilization, and HTTP errors -- shown as distinct node categories in the causal dependency graph?

Q2 (Module-level grouping). Are service boundaries clearly demarcated so that intra-service and inter-service dependencies are visually distinguished?

Q3 (Overall pipeline direction). Does the figure place the proposed instance-level method to the right of, or otherwise visually distinguish it from, the baseline aggregate-level approaches for direct comparison?
\end{promptbox}

\subsection{Rubric-Based Content Quality Questions}
\label{appendix:rubric_questions}

Rubric-Based Content Quality evaluates properties that are important across methodology figures, independent of the specific paper. Unlike C\&C, these questions are not tied to named entities in a particular method. Instead, the same 20 questions are applied to every generated figure, making the rubric scores directly comparable across papers, methods, and baselines. The rubric covers four aspects: technical accuracy, visual clarity, structural consistency, and readability.

We derive the rubric from 1{,}784 held-out high-quality methodology figures from the curation pipeline (Section~\ref{sec:data_curation}). These figures provide examples of recurring design conventions in published methodology diagrams, including consistent notation, clear module boundaries, traceable information flow, balanced visual density, and legible annotations. For each aspect, we curate 5 yes/no questions, yielding 20 questions total. Claude-Opus-4.6 answers each question using the generated figure and the corresponding methodology text. For each aspect, the score is the fraction of ``yes'' answers among its 5 questions, rescaled to the 0--10 scale used in Table~\ref{tab:main_results_rubric}. The mean of the four aspect scores gives the Rubric-Based Content Quality average.

Representative rubric questions are shown below; the full set of 20 questions is released with the code.

\begin{promptbox}{Technical accuracy rubric: notation, architecture, and transformations}
Technical Accuracy (Tech) -- example questions. Each question is binary yes/no.

T1. Does the figure use mathematical symbols, equations, and variable names consistently across all components, with no symbol changing meaning or font style between modules?

T2. Are the architectural components and connection types in the figure, such as skip connections, concatenations, and additions, consistent with the description in the methodology text?

T3. Are dimensional transformations, such as input/output shapes, projections, and aggregations, indicated explicitly at the points where the methodology text introduces them?
\end{promptbox}

\begin{promptbox}{Visual clarity rubric: hierarchy, density, and color use}
Visual Clarity (Clarity) -- example questions. Each question is binary yes/no.

C1. Does the figure use visual hierarchy, such as size, color, and position, to draw attention to the most important components first?

C2. Is the figure's information density balanced, neither overcrowded with detail nor so sparse that the pipeline is hard to follow?

C3. Is color used to convey meaningful information, such as distinguishing component types or grouping related modules, rather than purely decoratively?
\end{promptbox}

\begin{promptbox}{Structural consistency rubric: flow, boundaries, and grouping}
Structural Consistency (Cons) -- example questions. Each question is binary yes/no.

S1. Is the sequential flow of operations clearly indicated through directional arrows or unambiguous spatial ordering?

S2. Are the functional modules clearly delineated by visual boundaries, such as boxes, colored backgrounds, or consistent spacing?

S3. Are related components visually grouped into logical sub-modules in a way that reflects the method's hierarchical structure?
\end{promptbox}

\begin{promptbox}{Readability rubric: legibility, self-containment, and annotations}
Readability (Read) -- example questions. Each question is binary yes/no.

R1. Are the text labels, variable names, and annotations legible at the figure's intended publication size?

R2. Can the figure be understood from its own labels, legends, and visual cues without constant reference to the paper text?

R3. Are the legends, callouts, and supplementary annotations, if present, complete enough for a reader to interpret every symbol that appears in the figure?
\end{promptbox}

\subsection{Completeness Question Generation Prompt}
\label{appendix:cc_generation_prompt}

The per-paper C\&C question bank is produced by a single call to Claude-Opus-4.6. The prompt enforces two properties that make the C\&C score meaningful: every question must name a concrete entity from the paper, so that a generic-looking figure cannot answer it correctly by chance, and every question must be answerable from the rendered figure alone, so that the score reflects what the figure shows rather than what the paper text contains.

\begin{promptbox}{C\&C question-generation prompt: enforcing paper-grounded yes/no questions}
You are designing a Completeness and Correctness question bank for a specific scientific paper's methodology figure. The bank will be used to score generated figures of THIS paper via cross-model VQA.

You receive:
- the full methodology text of the paper,
- the original figure caption,
- the author-drawn reference figure.

Produce exactly 50 yes/no questions with ground-truth answers, distributed across the following 10 conceptual dimensions, with 5 questions per dimension:

1. Components present (named blocks, modules, sub-modules)
2. Named variables and tensors
3. Ordering of pipeline stages
4. Input/output correspondence between stages
5. Supervision signal and ground-truth flow
6. Loss terms and their attachment points
7. Module-level groupings (siamese branches, parallel paths, encoders)
8. Inter-module connections (arrows, skip connections, residuals)
9. Dimensional transformations (input/output shapes, projections)
10. Overall pipeline direction and topology

Rules:
- Every question MUST name a specific entity from the paper, such as a module name, variable, loss term, or explicit relationship.
- Generic visual-design questions are NOT allowed in this bank; they are covered by the separate Rubric-Based Content Quality bank.
- Every question MUST be answerable from the rendered figure alone.
- Each question is binary yes/no.
- Provide the ground-truth answer based on the author-drawn reference figure.

Output format:
**Q<i>:** <question text grounded in this paper's named entities>
**GT Answer:** <Yes/No>
**Dimension:** <one of the 10 dimensions above>
\end{promptbox}

\subsection{Scoring Protocol}
\label{appendix:scoring}

The two VLM-answered axes are reported separately in the main results table because they measure different properties of figure quality. C\&C measures paper-specific content recall, while Rubric-Based Content Quality measures general design and content-quality criteria. We therefore do not collapse them into a single aggregate score.

For C\&C, each of the three answering VLMs answers all 50 paper-specific questions for each generated figure using only the figure. We compute the answer accuracy against the ground truth for each VLM, average the three accuracies, and report the resulting score on the 0--10 scale used in Table~\ref{tab:main_results_rubric}. Per-method C\&C scores are means over the 435 benchmark papers.

For Rubric-Based Content Quality, Claude-Opus-4.6 answers the 20 fixed rubric questions for each generated figure using the figure and methodology text. For each of the four aspects (Tech, Clarity, Cons, Read), the aspect score is the fraction of ``yes'' answers among its 5 questions, rescaled to 0--10. The four aspect scores appear as separate columns in Table~\ref{tab:main_results_rubric}, and the Rubric-Based Content Quality average is their mean. Per-method scores are again averaged over the 435 benchmark papers.

All questions within a bank contribute equally. We do not apply per-question or per-aspect weights, and we do not weight individual judges by quality. Claude-Opus-4.6 generates the C\&C questions but is never used as a C\&C answering judge; conversely, Gemini-3.1-Pro, GPT-5.2, and Qwen3-VL answer C\&C questions but never generate them. Claude-Opus-4.6 does answer the rubric bank, but those 20 questions are derived from 1{,}784 real figures rather than generated by Claude-Opus-4.6 itself. Thus, no model is asked to score questions of its own making.

\section{Perceptual Design Quality Heuristics}
\label{appendix:perceptual_heuristics}

The Perceptual Design Quality (PDQ) axis measures low-level visual quality using deterministic image-processing heuristics rather than VLM judgments. This avoids sensitivity to judge model, prompt wording, and scoring scale, and makes the metric reproducible from the rendered figure alone. PDQ consists of eight metrics grouped into four categories: Color~\&~Contrast, Density~\&~Clutter, Layout~\&~Alignment, and Typography. Each metric returns a value in $[0,1]$, which we rescale to $0$--$10$ in Table~\ref{tab:main_results_heuristics}.

All metrics are computed from a shared figure representation containing the RGB image $I$, an estimated background color $\mathbf{b}$, a binary content mask $M$, connected-component bounding boxes $\mathcal{B}$, and text-like boxes $\mathcal{B}_{\mathrm{txt}}$. The content mask marks pixels that differ sufficiently from the background:
\begin{equation}
M_{ij}
=
\mathbb{1}\!\left[
\max_c |I_{ijc}-b_c| > 30
\right]
\label{eq:pdq_mask}
\end{equation}
We obtain $\mathcal{B}$ from connected components on $M$ and use $\mathcal{B}_{\mathrm{txt}}$ for typography-related metrics.

\paragraph{Color and Contrast.}
\emph{Access} measures the fraction of detected components whose foreground color satisfies the WCAG~2.1 AA contrast threshold~\cite{fernandez2019web21}. For each component $b\in\mathcal{B}$, we compute the foreground color $\mathbf{f}_b$ as the median color of its content pixels and evaluate
\begin{equation}
\mathrm{Access}
=
\frac{1}{|\mathcal{B}|}
\sum_{b\in\mathcal{B}}
\mathbb{1}\!\left[
\mathrm{CR}(\mathbf{f}_b,\mathbf{b}) \geq 4.5
\right]
\label{eq:pdq_access}
\end{equation}
where $\mathrm{CR}$ is the standard WCAG contrast ratio. \emph{Hue} measures palette coherence using Matsuda hue-harmony templates~\cite{matsuda1995color}. We extract dominant colors with $k$-means, discard achromatic colors, and score the best-fitting rotated hue template:
\begin{equation}
\mathrm{Hue}
=
\max_{\tau,\alpha}
\mathrm{cov}_{\tau}(\alpha)
\label{eq:pdq_hue}
\end{equation}
where $\tau$ indexes a harmony template and $\alpha$ its rotation.

\paragraph{Density and Clutter.}
\emph{Canvas} measures how much of the available canvas is used. Let $B^\star$ be the tight bounding box enclosing all content pixels in $M$. We define
\begin{equation}
\mathrm{Canvas}
=
\frac{|B^\star|}{H\cdot W}
\label{eq:pdq_canvas}
\end{equation}
This penalizes figures that occupy only a small portion of the canvas. \emph{Ink} approximates Tufte's data-ink ratio~\cite{tufte1983visualdisplay} by estimating the fraction of content pixels that carry semantic information rather than decorative fills or scaffolding lines:
\begin{equation}
\mathrm{Ink}
=
\frac{|M|-|D|}{|M|}
\label{eq:pdq_ink}
\end{equation}
where $D$ denotes pixels classified as decorative.

\paragraph{Layout and Alignment.}
\emph{Overlap} measures the fraction of detected component pairs whose bounding boxes do not intersect. Let $n=|\mathcal{B}|$ be the number of detected component boxes. For two boxes $b_i,b_j\in\mathcal{B}$, let $\mathrm{I}(b_i,b_j)$ denote their intersection area. We define
\begin{equation}
\mathrm{Overlap}
=
1
-
\frac{2}{n(n-1)}
\sum_{i<j}
\mathbb{1}\!\left[
\mathrm{I}(b_i,b_j)>0
\right],
\label{eq:pdq_overlap}
\end{equation}
with $\mathrm{Overlap}=1$ when $n<2$. Higher values indicate fewer unintended overlaps between detected components.
\emph{Uniform} measures whitespace balance by dividing the canvas into an $8\times8$ grid and computing the whitespace fraction $w_{ab}$ in each cell:
\begin{equation}
\mathrm{Uniform}
=
\max\!\left(
0,\,
1-
\frac{
\mathrm{std}\{w_{ab}\}_{a,b=1}^{8}
}{0.5}
\right)
\label{eq:pdq_uniform}
\end{equation}
High scores indicate that empty space is distributed evenly rather than concentrated in large blank regions.

\paragraph{Typography.}
\emph{Font} measures font-size consistency using the heights of text-like boxes in $\mathcal{B}_{\mathrm{txt}}$. We cluster text heights into approximate size tiers and penalize both within-tier variation and an excessive number of tiers:
\begin{equation}
\mathrm{Font}
=
\max\!\big(
0,\,
1-\overline{\mathrm{CV}}
-\rho\max(0,T-T_0)
\big)
\label{eq:pdq_font}
\end{equation}
where $\overline{\mathrm{CV}}$ is the average coefficient of variation within detected font-size tiers, $T$ is the number of tiers, $T_0=4$, and $\rho=0.05$. \emph{Text} measures font-style consistency using ink density inside text boxes as a proxy for stroke weight:
\begin{equation}
\mathrm{Text}
=
\max\!\left(
0,\,
1-
\mathrm{mean}_c\,\mathrm{CV}_c(\{d_b\})
\right)
\label{eq:pdq_text}
\end{equation}
where $d_b=|M\cap b|/|b|$ and $c$ indexes the detected font-size tiers.

\paragraph{Aggregation.}
Each category score is the mean of its two metrics, and the overall PDQ score is the mean of all eight:
\begin{equation}
\mathrm{PDQ}
=
\frac{1}{8}
\sum_{m\in\mathcal{M}_{\mathrm{PDQ}}}
m
\label{eq:pdq_aggregate}
\end{equation}
All metrics are equally weighted. Because PDQ depends only on deterministic image-processing operations, repeated evaluation of the same rendered figure produces the same score up to floating-point non-determinism.

\section{Data Curation and Human Studies}
\label{appendix:data_curation}

This section provides additional details on the final filtering stage in Section~\ref{sec:data_curation} and on the two human studies used in our paper: the binary keep/remove verification of candidate samples and the pairwise evaluation of generated figures.

\subsection{Text--Figure Consistency Filtering Prompt}
\label{appendix:filter_prompt}

After identifying candidate methodology sections and figures and applying the basic aspect-ratio and resolution filters, we evaluate each remaining (methodology text, figure) pair with Gemini-3.1-Pro and GPT-5.2. A sample is retained only if both models answer ``Yes,'' using a conservative AND rule that prioritizes precision over recall. This step reduces the corpus from 5{,}570 candidate papers to 2{,}632 filtered samples.

\begin{promptbox}{Text--figure consistency prompt: filtering candidate methodology figures}
You are checking whether a scientific figure faithfully depicts the methodology described in the accompanying text.

You are given:
- METHODOLOGY TEXT: the section of the paper that describes the proposed system.
- FIGURE: the candidate methodology figure from the same paper.

Decide whether the figure accurately depicts the proposed method, including its key components, modules, and overall information flow.

Respond with exactly one of:
- "Yes" -- the figure faithfully represents the method described in the text.
- "No" -- the figure omits, misrepresents, or contradicts core elements of the method, or depicts an unrelated diagram such as a results plot, teaser, or qualitative example.

Follow your answer with a one-sentence justification.
\end{promptbox}

\subsection{Human Verification Annotators}
\label{appendix:verification_annotators}

The final human verification stage was conducted by 60 AI/ML researchers recruited through university research-group mailing lists. Participants included undergraduate researchers, master's students, PhD students, and one postdoctoral researcher, with a median of 2 years of research experience (range: 0--10). Each candidate (text, figure, caption) sample was independently reviewed by two annotators, who made a binary keep/remove decision based on visual quality, readability, structural completeness, and alignment with the extracted methodology text. The two annotators agreed on 92\% of samples before adjudication; disagreements were resolved by a third annotator. This process yielded 2{,}219 valid samples, from which we selected the final 435-paper evaluation set and used the remaining samples for rubric development. Annotators provided informed consent and were compensated at the standard hourly rate for academic research assistance in their local region.

\subsection{Human Evaluation Volunteers}
\label{appendix:eval_volunteers}

The pairwise human evaluation in Table~\ref{tab:human_eval} used the same compensated 60-participant cohort. In each trial, participants saw two anonymized figures for the same paper, generated by two different methods, with left/right order randomized. They selected the better figure according to four criteria: technical accuracy, visual clarity, structural consistency, and readability. Participants could enlarge either figure for detailed inspection, and progress was saved automatically across sessions. In total, we collected 10{,}000 pairwise comparisons over 435 benchmark papers and 9 candidates per paper, comprising 8 generated systems and the author-drawn reference figures. We use these comparisons to compute the Elo ratings and pairwise win rates reported in Table~\ref{tab:human_eval}.

\end{document}